\documentclass[acmtog, nonacm]{acmart} 

\AtBeginDocument{%
  }

\usepackage{url,hyperref}
\usepackage{booktabs}

\usepackage{graphicx}
\usepackage{amsmath}
\usepackage{cleveref}
\usepackage{caption} 

\usepackage{amssymb}
\usepackage{bbm}
\usepackage{bm} 
\usepackage{booktabs}
\usepackage{multirow}
\usepackage{multicol}
\usepackage[table]{xcolor}

\usepackage[skip=6pt]{caption}

\usepackage{array}

\usepackage[ruled,vlined]{algorithm2e}

\usepackage{pifont}
\definecolor{limegreen}{HTML}{32CD32}
\definecolor{pinkl}{HTML}{FB2B97}

\def\ippn{LPPN}
\def\ippnfull{Local Pattern Producing Network}

\newcommand{\rev}[1]{#1}

\title{Neural Cellular Automata: From Cells to Pixels}

\begin{document}

\title{Neural Cellular Automata: From Cells to Pixels}


\author{Ehsan Pajouheshgar}
\email{ehsan.pajouheshgar@epfl.ch}
\affiliation{%
  \institution{EPFL}
  \country{Switzerland}
}
\author{Yitao Xu}
\affiliation{%
  \institution{EPFL}
  \country{Switzerland}
}

\author{Ali Abbasi}
\affiliation{%
  \institution{EPFL$^*$\thanks{$^*$Work done during internship at EPFL}}
  \country{Switzerland}
}

\author{Alexander Mordvintsev}
\affiliation{%
  \institution{Google Research}
  \country{Switzerland}
}

\author{Wenzel Jakob}
\affiliation{%
  \institution{EPFL}
  \country{Switzerland}
}

\author{Sabine Süsstrunk}
\affiliation{%
  \institution{EPFL}
  \country{Switzerland}
}



\begin{abstract}
    Neural Cellular Automata (NCAs) are bio-inspired dynamical systems in which identical cells iteratively apply a learned local update rule to self-organize into complex patterns, exhibiting regeneration, robustness, and spontaneous dynamics. Despite their success in texture synthesis and morphogenesis, NCAs remain largely confined to low-resolution outputs. This limitation stems from (1) training time and memory requirements that grow quadratically with grid size, (2) the strictly local propagation of information that impedes long-range cell communication, and (3) the heavy compute demands of real-time inference at high resolution. In this work, we overcome this limitation by pairing an NCA that evolves on a coarse grid with a lightweight implicit decoder that maps cell states and local coordinates to appearance attributes, enabling the same model to render outputs at arbitrary resolution. Moreover, because both the decoder and NCA updates are local, inference remains highly parallelizable. To supervise high-resolution outputs efficiently, we introduce task-specific losses for morphogenesis (growth from a seed) and texture synthesis with minimal additional memory and computation overhead. Our experiments across 2D/3D grids and mesh domains demonstrate that our hybrid models produce high-resolution outputs in real-time, and preserve the characteristic self-organizing behavior of NCAs.

\end{abstract}

\definecolor{pinkl}{HTML}{FB2B97}
\begin{teaserfigure}
  \centering
    \captionsetup{type=figure}
    \vspace{-10pt}
    \includegraphics[width=0.99\linewidth,trim={0 0 0 0},clip]{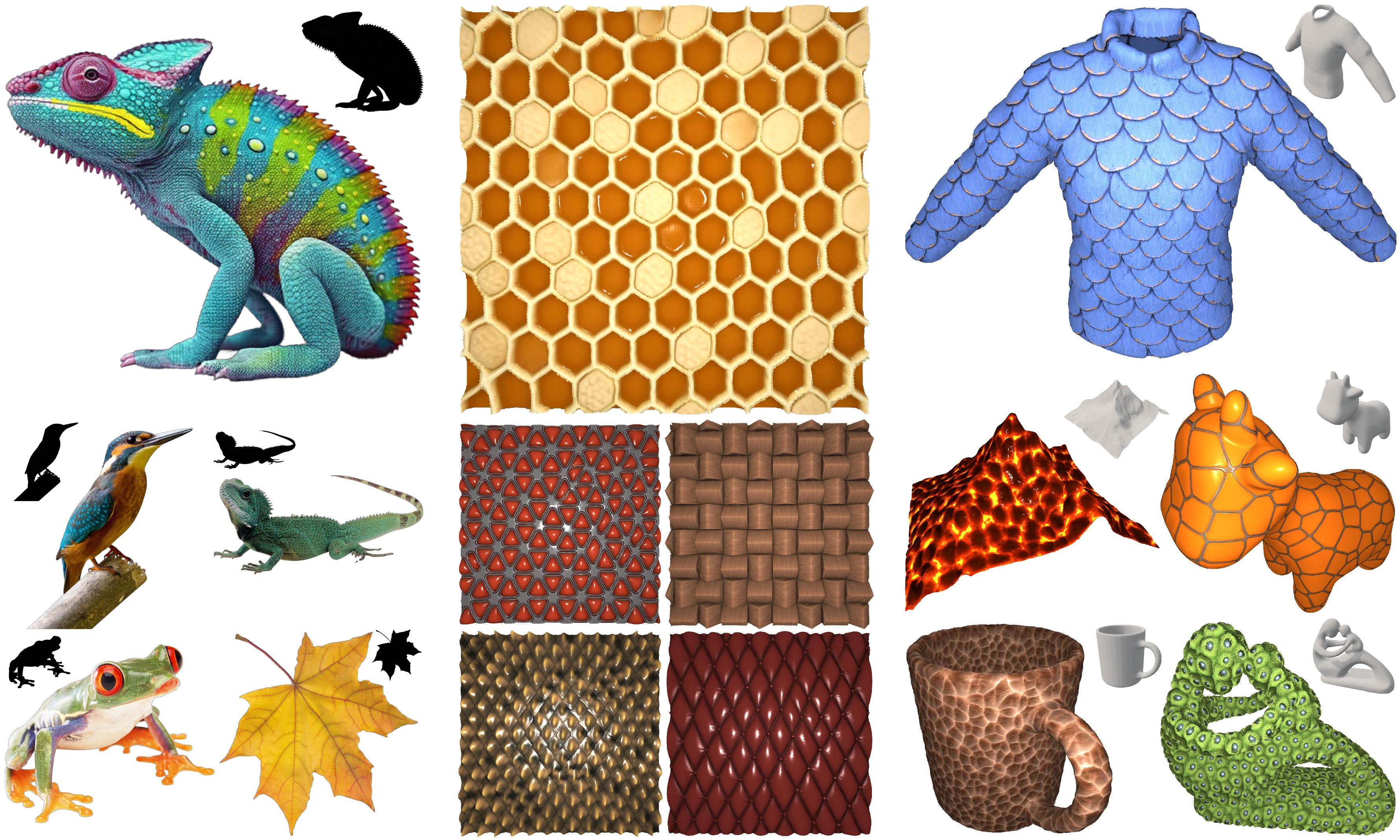}
    \vspace{-5pt}
    \captionof{figure}{Summary of results. Our proposed method enables NCAs to generate high-quality outputs with minimal extra cost. Our method is applicable to different NCA architectures and training targets. \textit{\textbf{Left}:} Growing 2D shapes and images from a single seed; \textit{\textbf{Middle}:} Texture synthesis in 2D; \textit{\textbf{Right}:} Texture synthesis on 3D Meshes. Online interactive demos are available at \href{https://cells2pixels.github.io}{\textcolor{pinkl}{https://cells2pixels.github.io}}.
    }
    \vspace{-0pt}
    \label{fig:teaser}
\end{teaserfigure}

\maketitle

\newcommand{\State}{\bm{s}}
\newcommand{\Output}{\bm{o}}
\newcommand{\Perception}{\mathrm{Z}}
\newcommand{\NState}{\mathrm{S}^{\odot}}

\newcommand{\CellSet}{\mathcal{C}}
\newcommand{\CellStateSet}{\mathcal{S}}
\newcommand{\PosSet}{\mathcal{P}}
\newcommand{\NeighborhoodSet}{\mathcal{N}}
\newcommand{\PerceptionStage}{\mathcal{Z}}
\newcommand{\AdaptationStage}{\mathcal{A}}

\newcommand{\Kx}{K_\textup{x}}
\newcommand{\Ky}{K_\textup{y}}
\newcommand{\Klap}{K_\textup{lap}}
\newcommand{\Kid}{K_\textup{id}}
\newcommand{\Wone}{\textup{W}_1}
\newcommand{\Wtwo}{\textup{W}_2}
\newcommand{\bone}{\textup{b}_1}
\newcommand{\Dx}{\Delta \textup{x}}
\newcommand{\Dy}{\Delta \textup{y}}
\newcommand{\Dt}{\Delta \textup{t}}
\newcommand{\Point}{\mathrm{\mathbf{p}}}

\vspace{-4pt}
\section{Introduction}

Complex self-organizing systems consist of numerous simple components that interact through simple\footnote{By simple rules, we mean interaction mechanisms with low descriptive complexity relative to the complexity of the resulting emergent behavior.} local\footnote{While locality of interactions is not strictly required in a complex system, every self-organizing system observed in nature relies on local interactions among its components.} rules, producing coherent macro behavior without centralized control. In nature, such systems manifest across many scales: elementary particles bind into atoms and molecules, which then assemble into the diverse materials we observe; a single fertilized cell undergoes differentiation to form a fully developed organism during morphogenesis; and neurons coordinate and synchronize their activations to support coherent cognitive functions. In all cases, the global structure emerges not from top-down planning but from the collective effect of countless simple local interactions.

Neural Cellular Automata (NCAs) make this principle learnable: a small neural update rule, shared across cells, can grow images and shapes from a seed \citep{mordvintsev2020growing,growing3d} and synthesize textures \citep{niklasson2021self-sothtml,dynca}.
Because NCAs learn an iterative self-organizing process rather than a direct mapping, they naturally exhibit robustness and regeneration \citep{mordvintsev2020growing}, generalize across discretizations and domains \citep{meshnca, noisenca}, and display emergent spontaneous motion \cite{emergent_dynamics}, while encoding this process with a lightweight, compute-efficient neural network.

In practice, however, current NCAs are trained on grids, or meshes with at most $10^4\!\sim\!10^5$ cells \citep{mordvintsev2020growing,meshnca}, which translates to spatial resolutions of about $64\times64$ to $256\times256$ depending on the space dimensionality.
Scaling further is difficult: memory and compute grow quickly with the number of cells, while information still propagates locally (one neighborhood hop per update), requiring more steps for long-range coordination.
As a result, it remains unclear how to obtain high-resolution, high-quality outputs without abandoning locality or efficiency.

\begin{figure}[!t]
    \centering
    \vspace{-5pt}
    \includegraphics[width=\linewidth]{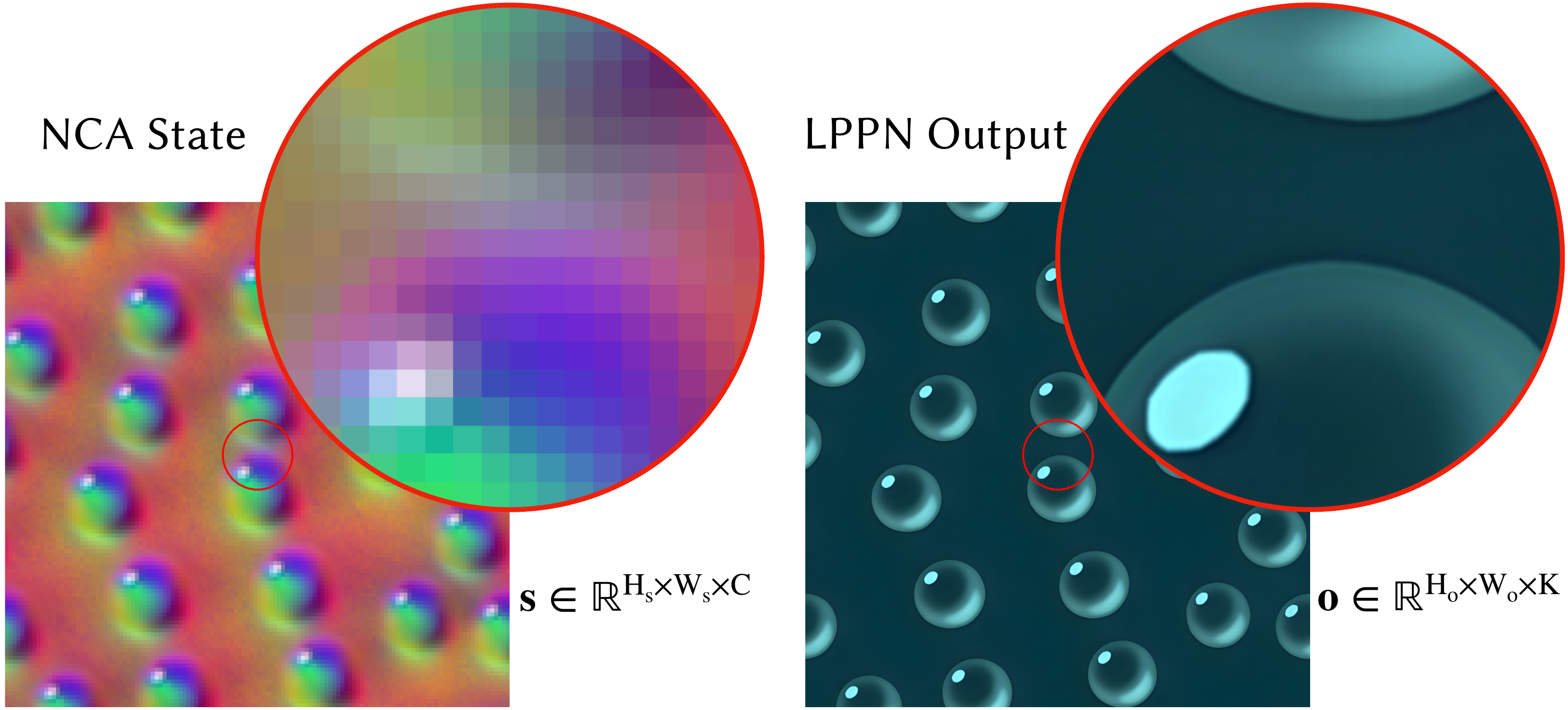}
    \vspace{-12pt}
    \caption{
    \textbf{Sample Output of Our Hybrid Model.}
    The NCA evolves on a coarse $128 \times 128$ lattice (left); 
    Our \ippnfull{} (\ippn{}) acts as a neural field decoder that decouples the render resolution from the NCA grid size.
We sample this field at $1024 \times 1024$ and, without retraining, at $8192 \times 8192$ in the magnified inset (right).
    }
    \vspace{-12pt}
    \label{fig:nca_ippn_output}
\end{figure}

We address this limitation by decoupling \emph{dynamics} from \emph{appearance}.
We evolve the NCA on a coarse lattice, but render a continuous, high-resolution output with a lightweight coordinate-based decoder, \ippn{}, which acts as a neural field conditioned on the local NCA state\footnote{From the biology perspective, the NCA provides the high-level blueprint that determines where structure should emerge, while the \ippn{} sidesteps the intricate biochemical refinement processes and instead renders those details in a single pass.}.
Concretely, \ippn{} maps a locally interpolated cell feature and an intra-primitive coordinate to appearance at arbitrary query locations, decoupling the render resolution from the NCA lattice size as shown in Figure~\ref{fig:nca_ippn_output}.
We train the NCA and \ippn{} end-to-end; \ippn{} adds only $20\!\sim\!30\%$ extra parameters. Since all recurrent updates happen at low resolution\footnote{This mirrors the common strategy in modern vision models (e.g., transformers or latent diffusion) of performing the expensive computation on a compact set of tokens/latents.}, training remains memory-efficient and fast, and inference remains real time.

We evaluate our hybrid model across four settings: morphogenesis, 2D PBR texture synthesis, texture synthesis on meshes, and 3D volumetric texture synthesis.
Across domains, we show that \ippn{} improves output quality at high resolution with minimal overhead, while retaining the characteristic self-organizing properties of NCAs, along with their efficiency and interactive controllability.
We also provide an interactive web demo (\href{https://cells2pixels.github.io}{\textcolor{pinkl}{\nolinkurl{cells2pixels.github.io}}}) that runs our trained models fully on-device in the browser. 

\vspace{-5pt}
\section{Related Works}
\vspace{-2pt}
\subsection{Neural Cellular Automata}
\vspace{-1pt}

Early computational explorations of self-organization trace back to Alan Turing's reaction-diffusion model of morphogenesis \cite{turing-pattern} and John von Neumann's formulation of cellular automata \cite{vonneumann-introca}. Both lines of work revealed that simple hand-crafted local rules, applied over time, can yield intricate global patterns. Carefully hand-designed CA and reaction-diffusion rules have reproduced 2D and 3D morphogenesis and texture phenomena \cite{cellular-texture-generation,3d-surface-ca,turk-rd-texture}, yet the need for laborious manual rule search has long limited further exploration. 
Renewed attention to cellular automata began with \citet{gilpin2019cellular}, which showed that a classical CA update can be written as a shallow convolutional neural network and trained with back-propagation.
Extending this idea, \citet{mordvintsev2020growing} parameterize the update rule by an MLP, and use fixed convolutional kernels for cell interaction modeling, giving rise to NCA. 

\begin{figure*}[!htbp]
    \centering
    \includegraphics[width=\linewidth, trim={5pt 0pt 0pt 0pt}, clip]{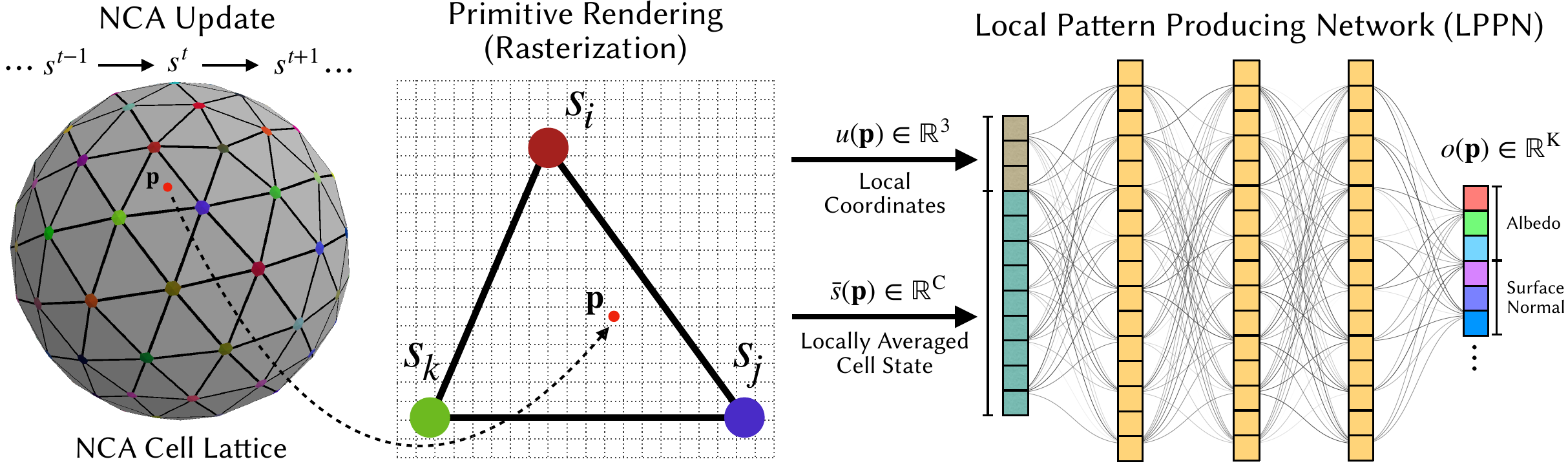}
    \vspace{-12pt}
    \caption{
    \textbf{Hybrid NCA\,+\,\ippn{} Overview.}
    \textit{\textbf{Left}:} The NCA operates on a coarse lattice of cells (in this example vertices of a mesh)
    \textit{\textbf{Center}:} A sampling point $\Point$ (red dot) inside a triangle primitive, whose vertices correspond to NCA cells $\State_i,\,\State_j,\,\State_k$.  
    The \emph{local coordinate} $u(\Point)$ expresses the point's position inside the primitive, while the \emph{locally averaged cell state} $\bar{\State}(\Point)$ is obtained by interpolating the surrounding cell states.    
    \textit{\textbf{Right}:} A shared lightweight MLP, \ippn{}, receives $(\bar{\State}(\Point),u(\Point))$ as input and outputs the appearance features, such as color and surface normal, at point $\Point$.}
    \label{fig:pipeline}
    \vspace{-10pt}
\end{figure*}

NCAs have been used to model morphogenesis: \citet{mordvintsev2020growing} grows 2D images from a seed while demonstrating regenerative capabilities, \citet{growing3d} extends the idea to 3D voxel grids by using 3D convolutions, and \citet{goal-nca} conditions the update rule to grow multiple targets.
NCAs have also been successful for texture synthesis: \citet{niklasson2021self-sothtml} demonstrated exemplar-based texture synthesis with emergent motion, \citet{dynca} added controllable dynamics, \citet{noisenca} improved stability and scale control via noise initialization, \citet{meshnca} generalized NCAs to meshes via non-parametric message passing, and \citet{wang2025volumetric,mokume} extend this framework to 3D to synthesize volumetric textures. Beyond morphogenesis and texture synthesis, NCAs have also been used for generative modeling~\citep{vnca,gannca,selfattentionnca,kalkhof2024frequency} and discriminative tasks such as classification, segmentation, and robust representation learning~\citep{randazzo2020self-classifying,adanca,nca-imageseg,mednca,arcnca}.

Despite this progress, most NCA models still operate on relatively modest cell counts, which limits the achievable output resolution and detail. Scaling further is impeded by GPU memory limits, slow information propagation in large grids, and the locality-breaking global operations often introduced as work-arounds.
We overcome this limitation by keeping the NCA coarse while using a lightweight, local neural field decoder to render fine detail at arbitrary resolution.

\vspace{-5pt}
\subsection{Neural Fields}
\vspace{-2pt}

Neural fields (a.k.a.\ implicit neural representations) model signals as continuous coordinate-based functions \citep{cppn}.
They have become central in vision and graphics, e.g., DeepSDF for shape fields \citep{park2019deepsdf}, NeRF for view-dependent radiance and density \citep{nerf}, and SIREN for high-frequency signal fitting via sinusoidal activations \citep{sitzmann2019siren}.
While neural fields are resolution-free, semantic structure is typically embedded in dense weights, making them harder to interpret and manipulate.

Our hybrid design combines the strengths of both paradigms: the NCA produces an explicit, local, editable lattice of features, while \ippn{} uses these features as conditioning to render a continuous high-resolution field.
This preserves locality and interactivity while lifting the resolution bottleneck.

\vspace{-5pt}
\section{Method}
\vspace{-2pt}
\subsection{Preliminaries - Neural Cellular Automata}

Neural Cellular Automata (NCAs) models the temporal evolution of a set of cells on a grid. Let $\State_{i,t} \in \mathbb{R}^C$ denote the $C$-dimensional vector storing the $i$-th cell state at time $t$. At each time step, cells perceive their neighbors in the \textbf{Perception} stage $\PerceptionStage{}$, to collect the information of their local neighborhood. Based on this information, cells adapt to the environment by updating the states in the \textbf{Adaptation} stage $\AdaptationStage{}$. The perception $\PerceptionStage{}$ is often instantiated using convolution for 2D and 3D grids \cite{mordvintsev2020growing,niklasson2021self-sothtml,dynca,noisenca, growing3d}, or a convolution-like operator for meshes \cite{meshnca}. The adaptation function $\AdaptationStage{}$ is commonly parameterized by a Multi Layered Perceptron (MLP), which sometimes includes a stochastic term \cite{mordvintsev2020growing,niklasson2021self-sothtml,dynca,meshnca,adanca}.
The state of cell $i$, $\State_{i}$, is updated as in Equation.~\ref{eq;nca-discrete}. 
\vspace{-2pt}
\begin{equation}
    \State_{i}^{t+\Delta t}=\State_{i}^{t} + \AdaptationStage{}(\PerceptionStage{}(\State_{i}^{t},\bm{s}_{j}^{t})) \cdot \Delta t, \quad j \in \NeighborhoodSet{}(i).
    \label{eq;nca-discrete}
\end{equation}
\vspace{-2pt}
$\NeighborhoodSet{}(i)$ denotes the set of neighbors of cell $i$. 
After evolving for $T$ steps, the cell states $\bm{s}_T$ of all cells can be extracted for further downstream applications. 
We forward the resulting cell states to the \ippnfull{}, which transforms the low-resolution NCA features in $\bm{s}_T$ into high-resolution detailed output.

\vspace{-4pt}
\subsection{\ippnfull{} (\ippn{})}
\label{sec:ippn}

The \ippn{} converts the coarse NCA lattice into a high-resolution field by evaluating a small MLP decoder network at every sampling point $\Point$, as shown in Figure~\ref{fig:pipeline}.
\rev{A sampling point may be the center of a rasterized pixel for 2D images or 3D meshes, or a 3D position for a NeRF-style volumetric rendering.}
For each $\Point$, the decoder receives:

\begin{enumerate}
    \item \emph{locally interpolated cell state} $\bar{\State}(\Point)\in\mathbb{R}^{C}$, which aggregates the states of the cells in $\Point$'s surrounding \emph{primitive}. 
    \item \emph{local coordinate} vector $\bm{u}(\Point)$ that encodes the relative position of $\Point$ inside its \emph{primitive}.
\end{enumerate}

Below, we define the primitive associated with sampling points and define locally interpolated state and local coordinates.


\begin{figure*}[!ht]
    \centering
    \includegraphics[width=\linewidth]{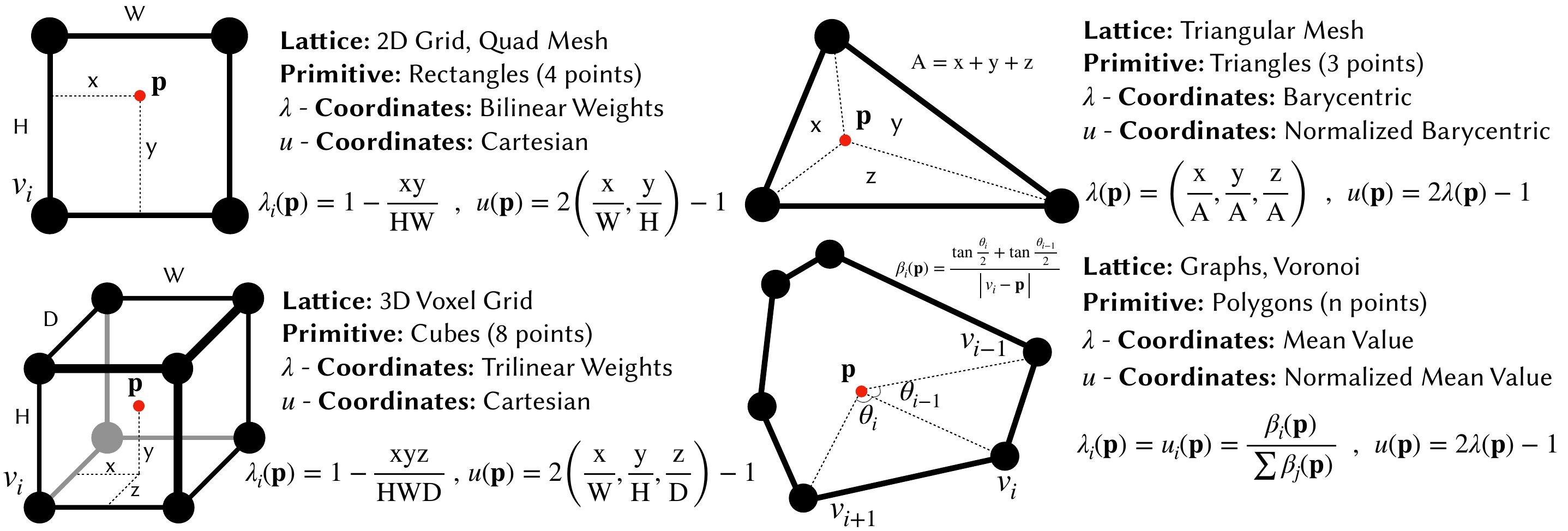}
    \vspace{-10pt}
    \caption{\rev{Representative examples of primitives. Vertices correspond to neighboring cells of an arbitrary sampling point $\Point$.}}
    \label{fig:primitives}
    \vspace{-8pt}
\end{figure*}

\vspace{-3pt}
\subsubsection{Primitives}
A \emph{primitive} $\Omega$ consists of the cells surrounding a sampling point $\Point$ along with the geometric shape determined by their positions.
A primitive's shape is determined by the underlying lattice; on a 2D quad grid, for instance, the de facto primitive would be a rectangle.  
Figure~\ref{fig:primitives} illustrates four common primitive types.
For $i \in \Omega$, let $v_i$ denote the position of the $i$-th cell in the primitive. 
We call a set of weights $\lambda_{i}(\Point)$ that describe the position of any arbitrary $\Point$ inside the primitive with respect to the primitive's vertices a $\lambda$-coordinate if it satisfies the following three conditions:

\begin{enumerate}
    \item \textbf{Partition of unity:} $\sum_{j \in \Omega} \lambda_j(\mathrm{\mathbf{p}})=1$.
    \item \textbf{Non-negativity} $\lambda_j(\mathrm{\mathbf{p}}) \geq 0$.
    \item \textbf{Linear precision:} $\mathrm{\mathbf{p}}=\sum_{j \in \Omega} \lambda_j(\mathrm{\mathbf{p}}) v_j$.
\end{enumerate}

While triangles admit a unique $\lambda$-coordinate (barycentric), higher-order primitives offer multiple valid choices.
For the rectangular primitive, a valid choice for $\lambda$-coordinates  is the bilinear interpolation weights. Figure~\ref{fig:primitives} illustrates the most common variants adapted in this paper for four different types of primitives.
With the chosen $\lambda$-coordinate system in place, we obtain the first \ippn{} input by averaging the cell states within $\Point$'s primitive $\Omega$:

\begin{equation}
    \bar{\State}(\mathrm{\mathbf{p}})=\sum_{j\in \Omega} \lambda_j(\mathrm{\mathbf{p}})\State_j
\label{eq:ippn-avg}
\end{equation}

\subsubsection{Local Coordinates}

Rather than feeding the full $\lambda$-coordinates directly to the \ippn{}, we transform them to a more compact \emph{local coordinate} $u(\Point)$, which is easier for the decoder to digest.
For rectangle and cube primitives we use axis-aligned Cartesian coordinates for $u$. This representation fully determines the point's location inside the primitive with fewer dimensions than the eight (or four) dimensional $\lambda$-coordinates.
For triangular meshes and general polygonal primitives we retain the $\lambda$-coordinates. In every case we rescale each component so that $u(\Point)\in[-1,1]$, whereas the original $\lambda$-coordinates lie in $[0,1]$. This zero-centered range removes a potential bias in the decoder's inputs and empirically improves learning.  
Figure~\ref{fig:primitives} summarizes the specific choices of $\lambda$-, and $u$-coordinates for the primitive types considered in this paper. Before passing $u(\Point)$ to the \ippn{}, we optionally apply simple transformations that improve its continuity and uniformity, as detailed next.

\begin{figure}[!htbp]
\centering
\includegraphics[width=\linewidth]{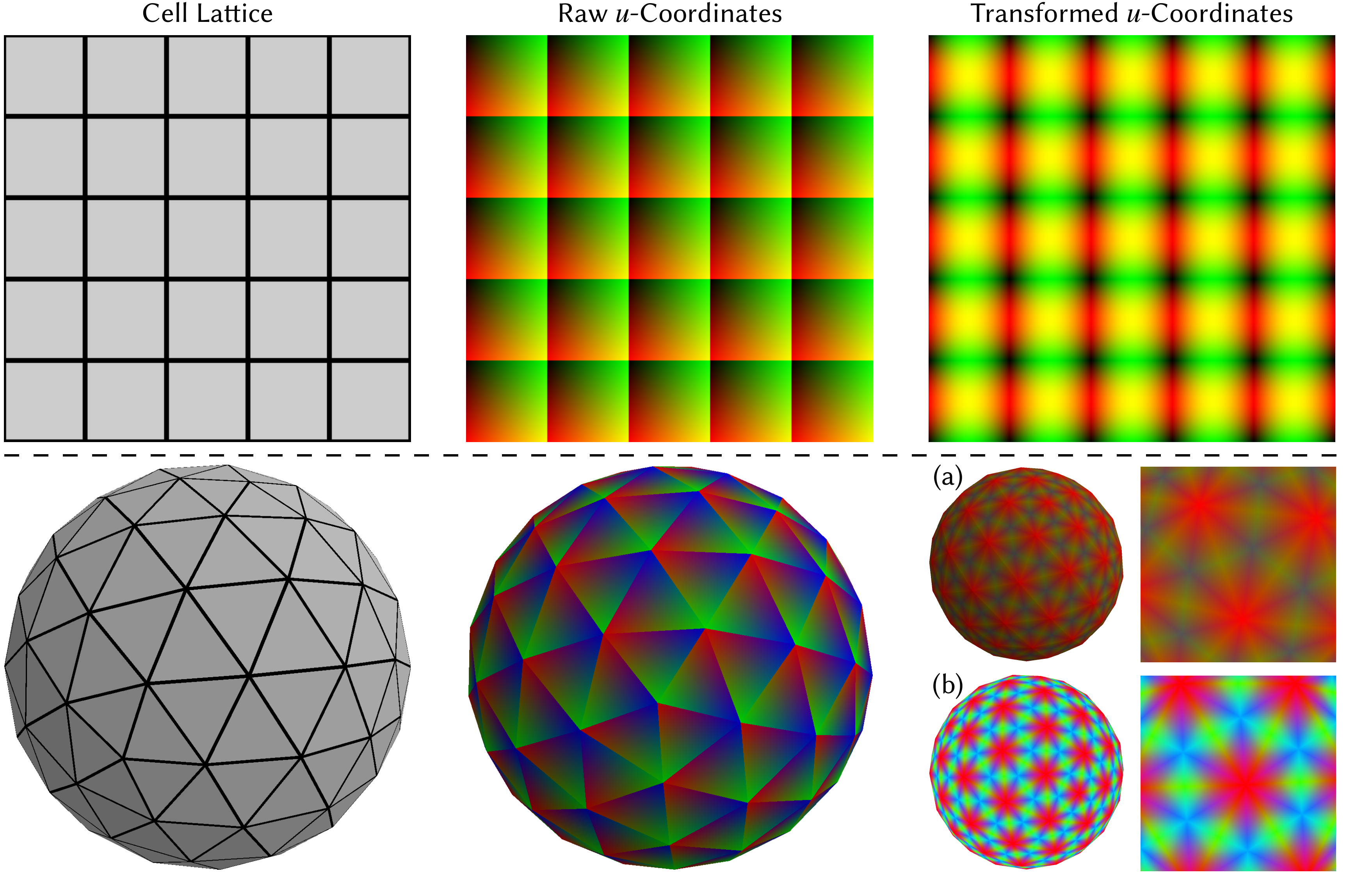}
\vspace{-9pt}
\caption{\textbf{Local coordinate transformations.}
Raw $u$-coordinates visualized as RGB suffer discontinuities at primitive boundaries. (Top Right) Applying trigonometric functions to Cartesian coordinates enforces $C^{0}$ continuity at boundaries for rectangle primitives. 
\emph{Bottom Right (a): } Sorting the barycentric coordinates enforces $C^{0}$ continuity but yields an imbalanced dynamic range (red color dominates).
\textit{(b)} Applying a remapping equalizes the range, giving a uniform, continuous positional field that is easier for the \ippn{} to digest.}
\vspace{-9pt}
\label{fig:coord_transformation}
\end{figure}

\vspace{-3pt}
\subsubsection{Ensuring $C^0$ Continuity and Uniformity}

Raw $u$-coordinates are piecewise smooth yet generally \emph{discontinuous} at primitive boundaries, as shown in the middle column of Figure~\ref{fig:coord_transformation}. 
On a 2D grid (top), the Cartesian coordinates reset at each primitive, and on a triangular mesh (bottom) the barycentric coordinates change abruptly across edges. 
To provide the \ippn{} with smooth, well-conditioned positional cues, we apply simple, primitive-specific transforms to the raw $u$-coordinates.
For rectangle and cube primitives we keep the Cartesian coordinates but encode them with a sinusoidal basis of the first $n$ harmonics,
\begin{equation}
\bm{u}_{\text{aug}}
  = \!\bigl[\,\sin(\pi \bm{u}),\;\cos(\pi \bm{u}),\;
             \dots,\;
             \sin(n\pi \bm{u}),\;\cos(n\pi \bm{u})\,\bigr].
\end{equation}
This encoding is continuous across primitive boundaries (Fig.~\ref{fig:coord_transformation}, top-right) and is injective on the interior of each primitive.

For triangular or polygonal primitives, the raw $u$-coordinates are discontinuous at primitive boundaries because neighboring faces list their vertices in arbitrary orders. To eliminate this order-dependence we first sort the $u$-coordinates (barycentric/mean-value) in descending order, making the largest weight always the first component. This consistent reordering makes the local coordinate field $C^{0}$ across primitive boundaries, as illustrated in Figure \ref{fig:coord_transformation},(a). Sorting, however, skews the dynamic range of the different components of $u$ (visible as the predominance of red in Figure~\ref{fig:coord_transformation}(a)).
To better balance these inputs and make it easier for the \ippn{} to utilize all coordinates, we apply a simple monotone remapping to each sorted component so that its values are spread more uniformly over $[-1,1]$; see Appendix for details. The resulting coordinates have comparable amplitude across components and remain continuous across primitives (Figure~\ref{fig:coord_transformation}(b)). Although sorting makes the coordinates non-injective, \ippn{} also conditions on $\bar{\State}(\Point)$, which provides the missing context and makes the joint input expressive enough for high-quality decoding.



\subsubsection{MLP Decoder}
The \ippn{} is a lightweight MLP $\bm{f}_{\theta}$ that takes the locally averaged state and local coordinates of each sampling point $(\bar{\State}(\Point),u(\Point))$, and returns the desired attributes at the sampling point:
\begin{equation}
\Output(\Point)=\bm{f}_\theta \;\bigl(\bar{\State}(\Point),u_{\text{aug}}(\Point)\bigr)\in\mathbb{R}^{K},
\label{eq:ippn-output}
\end{equation}

where $K$ is the number of channels in the output, e.g. 
$K=3$ for RGB-only, or $K=9$ when the base color is augmented with surface normals, height, ambient occlusion, and roughness maps.
Because \ippn{} is evaluated only during loss computation, all recurrent NCA updates still run on the coarse lattice. Therefore adding the \ippn{} incurs negligible training overhead.

\vspace{-4pt}
\section{Loss Functions}
We design task-specific objectives for our two main settings: texture synthesis and morphology growth from a seed.

\vspace{-4pt}
\subsection{Texture Loss}
\label{sec:texture-loss}
We supervise texture synthesis with a VGG-based texture loss built on the relaxed optimal transport (OT) style loss of \citet{kolkin2019style-otloss}, as adopted in prior NCA texture work~\cite{dynca,meshnca}. Concretely, we extract features from a fixed pretrained VGG16~\cite{vgg} at a few layers and match the distributions of generated and target features using OT with additional moment matching (mean and covariance)~\cite{gatys2015texture}. We extend this baseline with (i) patch-based multi-scale supervision, (ii) pseudo targets for multi-map PBR texture alignment, and (iii) an auto-correlation regularizer. All losses are computed on images rendered from the \ippn{} output.

\textbf{Patch-based multi-scale supervision.}
Prior work applies VGG on an image pyramid to capture texture statistics across multiple scales~\cite{snelgrove2017high,gonthier2022high}, but running VGG on full-resolution pyramids becomes increasingly expensive for high-resolution outputs. We therefore evaluate the OT style loss at three scales (full, half, and quarter target resolution) using random \emph{patches} from the rendered output: at each scale we take random crops with the number and size of the crops increasing with the scale, resize each crop to a fixed input size (e.g., $256\times256$), and compute the OT style loss. On the target side, we do \emph{not} crop so as not to lose any information; instead we build a target pyramid by resizing the exemplar per scale and caching its VGG features. This patch-based multi-scale loss avoids the quadratic blow-up of feeding full-resolution pyramids to VGG and the cost scales primarily with the number of crops. 
Moreover, it is particularly well-suited to our setting as both the NCA update rule and the \ippn{} decoder are local and weight-shared, making the gradients computed from random crops representative of the entire output.

\textbf{Pseudo targets for PBR map alignment.}
For PBR texture synthesis, the \ippn{} outputs 9 values corresponding to three maps: albedo, normal, and HRA (height, roughness, ambient occlusion). Applying the texture loss independently per map can lead to cross-map misalignment~\cite{meshnca}. To explicitly couple different texture maps, we add \emph{pseudo targets}: at each iteration we randomly select three distinct channels from the union of all target maps, stack them into a synthetic 3-channel target, and apply the same multi-scale OT loss to the corresponding three generated channels. This channel mixing provides an explicit alignment signal across maps (validated in our ablations).

\textbf{Auto-correlation regularizer.}
To encourage long-range geometric structure, we optionally add an FFT-based auto-correlation loss on early VGG feature maps, following correlation-based texture constraints~\cite{sendik2017deep,gonthier2022high}. We compute a channel-aggregated auto-correlation map for target and generated features and penalize their L1 difference; in practice, applying this term only at the \emph{coarsest} scale of our multi-scale loss is sufficient. We enable this term only for textures with strong geometric structure and long-range correlations.

\vspace{-4pt}
\subsection{Morphology Loss}
\label{sec:morphology-loss}
We train an NCA to grow a target morphology from a single seed, following \citet{mordvintsev2020growing}. The target is an RGBA image whose alpha channel defines the desired shape; We pad the target into a larger transparent canvas to give the morphology room to expand. In our hybrid model, the NCA evolves on a coarse lattice and produces a \emph{living mask} (one channel) that gates updates, while \ippn{} decodes the NCA state into a high-resolution RGBA image. Our morphology loss is the (unweighted) sum of:


\textbf{(i) RGBA reconstruction on \ippn{} output.}
We apply \rev{a combination of pixel-wise L1 and L2 losses} between the predicted and target RGBA images after masking both the generated output and the target image by their corresponding alpha channels.


\textbf{(ii) Shape loss on the NCA living mask.}
Since only living cells (and their neighbors) can update their state, directly supervising the living mask is essential to teach the NCA to expand from the seed. We bilinearly upsample the living mask to the render resolution and apply \rev{pixel-wise L1 and L2 losses} against the target alpha channel.


\textbf{(iii) LPIPS for sharp outputs.}
Because growth is stochastic and local, small variations and slight misalignments are common; optimizing only \rev{pixel-wise L1 and L2 losses} tends to average over plausible outcomes and produce blurry outputs. To counteract this, we add an LPIPS loss \cite{lpips} to encourage the model to generate perceptually similar texture and appearance to the target. Since LPIPS expects three channels, we compute it by randomly selecting three of the four RGBA channels per batch element.

\rev{The locality of the update rule together with the shape loss encourage the learned dynamics to naturally follow the silhouette of the target morphology during the growth process.}

\vspace{-4pt}
\section{Experiments}
\label{sec:experiments}

We evaluate our hybrid NCA+\ippn{} framework on four representative settings:
(i) \textbf{Growing a Morphology},
(ii) \textbf{Synthesizing PBR Textures},
(iii) \textbf{Synthesizing Textures on Meshes}, and
(iv) \textbf{Synthesizing 3D Volumetric Textures}.
Table~\ref{tab:exp_configs} summarizes the architecture and training hyperparameters for each experiment. 

We follow standard NCA training practices (checkpoint pool, stochastic updates, random rollout lengths, overflow regularization, and gradient normalization).
In all experiments we use a 4-layer SIREN MLP~\citep{sitzmann2019siren} as \ippn{} (three $\sin$ layers + linear output), adding $~\approx 25\%$ parameters relative to the NCA.


Our primary baseline replaces \ippn{} with a single linear readout from the locally averaged cell state (no coordinate input); matching the baseline parameter budget by increasing NCA channels yields similar results, so we report the simpler baseline with matched NCA parameter count.
We do not compare to an NCA operating directly at render resolution, as its training memory requirements exceed current GPU memory limits. Full training details for each experiment are provided in the Appendix. 
\textbf{All figures are best viewed digitally after zooming. }


\begin{table}[]
\centering
\caption{\textbf{Experiment configurations.}
We report NCA/\ippn{} architecture and training hyperparameters.
$^\dagger$Input dim $=$ locally averaged NCA state $+$ local coordinate encoding.}
\vspace{-3pt}
\label{tab:exp_configs}
\renewcommand{\arraystretch}{1.08}
\scriptsize
\resizebox{\columnwidth}{!}{%
\begin{tabular}{llcccc}
\toprule
\multicolumn{2}{c}{Setting} &
\multicolumn{1}{c}{Growing} &
\multicolumn{1}{c}{PBR-2D} &
\multicolumn{1}{c}{Mesh} &
\multicolumn{1}{c}{Vol-3D} \\
\midrule

\multirow{5}{*}{NCA}
& Domain
& 2D grid & 2D grid & Icosphere & 3D grid \\
& \#Cells
& $96^2$ & $128^2$ & 40k & $64^3$ \\
& Channels
& 32 & 16 & 16 & 16 \\
& MLP Width
& 256 & 128 & 128 & 128 \\
& \#Params
& 41k & 10k & 12k & 12k \\
\midrule

\multirow{4}{*}{\ippn{}}
& Input dim
& $32{+}4$ & $16{+}2$ & $16{+}3$ & $16{+}3$ \\
& Output dim ($K$)
& 4 & 9 & 3 & 4 \\
& MLP Width
& 64 & 32 & 32 & 32 \\
& \#Params
& 11k & 3k & 3k & 3k \\
\midrule

\multicolumn{2}{l}{Output Resolution}
& $768^2$ & $1024^2$ & $1024^2$ & $512^2$ \\
\multicolumn{2}{l}{Train Iterations}
& 20k & 3k & 3k & 4k \\
\multicolumn{2}{l}{Batch Size}
& 8 & 2 & 2 & 4 \\
\multicolumn{2}{l}{Rollout Steps}
& $[32,96]$ & $[32,128]$ & $[32,96]$ & $[16,48]$ \\
\bottomrule
\end{tabular}%
}
\vspace{-15pt}
\end{table}

\vspace{-4pt}
\subsection{Growing a Morphology}
\label{sec:exp_morphology}

We train an NCA to grow a target morphology from a single seed.
Targets are provided at $512{\times}512$ and embedded in a $768{\times}768$ transparent canvas to give the morphology room to expand.
The NCA evolves on a $96{\times}96$ grid and exposes a \emph{living} channel that gates updates; \ippn{} decodes the final state into a $768{\times}768$ RGBA image supervised by the morphology loss (Sec.~\ref{sec:morphology-loss}).
We encode local 2D coordinates with $\sin/\cos$ pairs to enforce $C^0$ continuity (Sec.~\ref{sec:ippn}).
To improve robustness to poor local minima, we periodically \rev{reset the pool to seed states} and restart the learning-rate schedule.

\begin{figure}[]
\centering
\includegraphics[width=\linewidth]{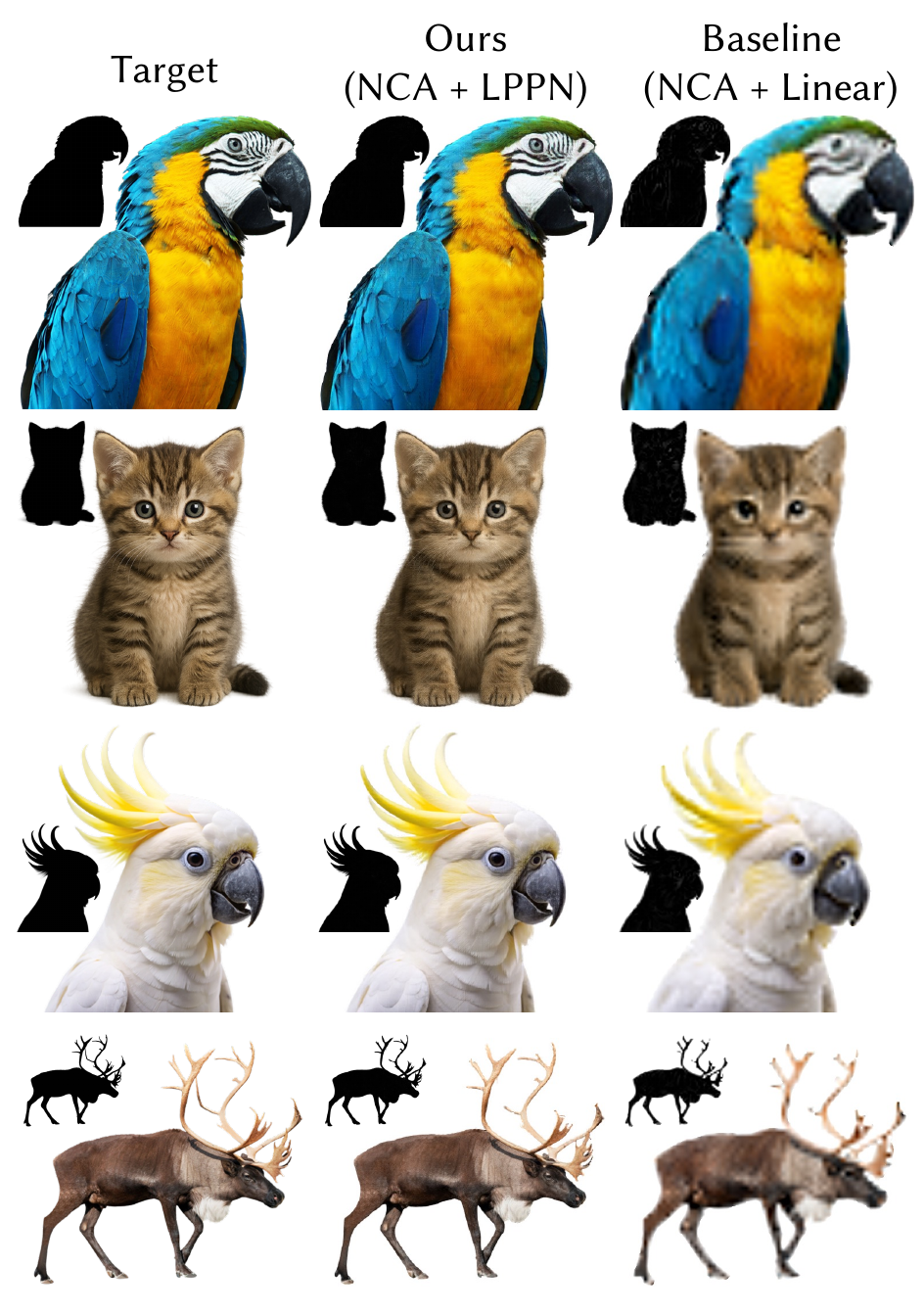}
\vspace{-20pt}
\caption{\rev{Morphology growth: comparison to the baseline.}}
\vspace{-13pt}
\label{fig:morphology_qual}
\end{figure}

\begin{figure}[]
\centering
\includegraphics[width=\linewidth]{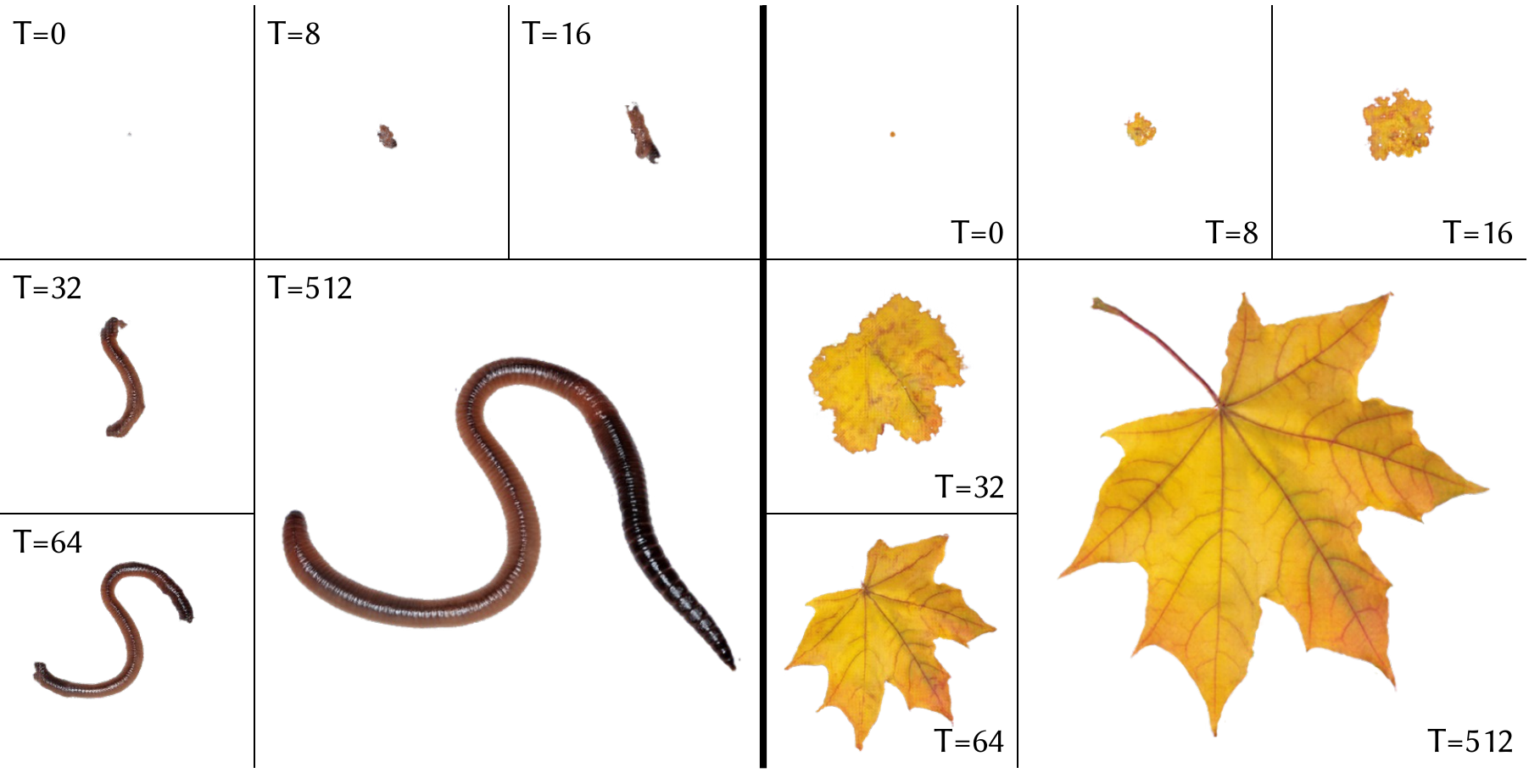}
\vspace{-10pt}
\caption{NCA growth from a single seed to the target morphology over time.}
\vspace{-5pt}
\label{fig:morphology_growth}
\end{figure}

Figure~\ref{fig:morphology_qual} compares our method to the primary baseline.
Our outputs have consistently sharper shapes and structures with finer appearance details, while the baseline tends to blur boundaries and wash out high-frequency content.
Figure~\ref{fig:morphology_growth} visualizes the evolution over time: starting from a single seed, the learned dynamics expand outward and naturally grow the target silhouette and morphology.

\begin{figure}[]
    \centering
    \includegraphics[width=\linewidth]{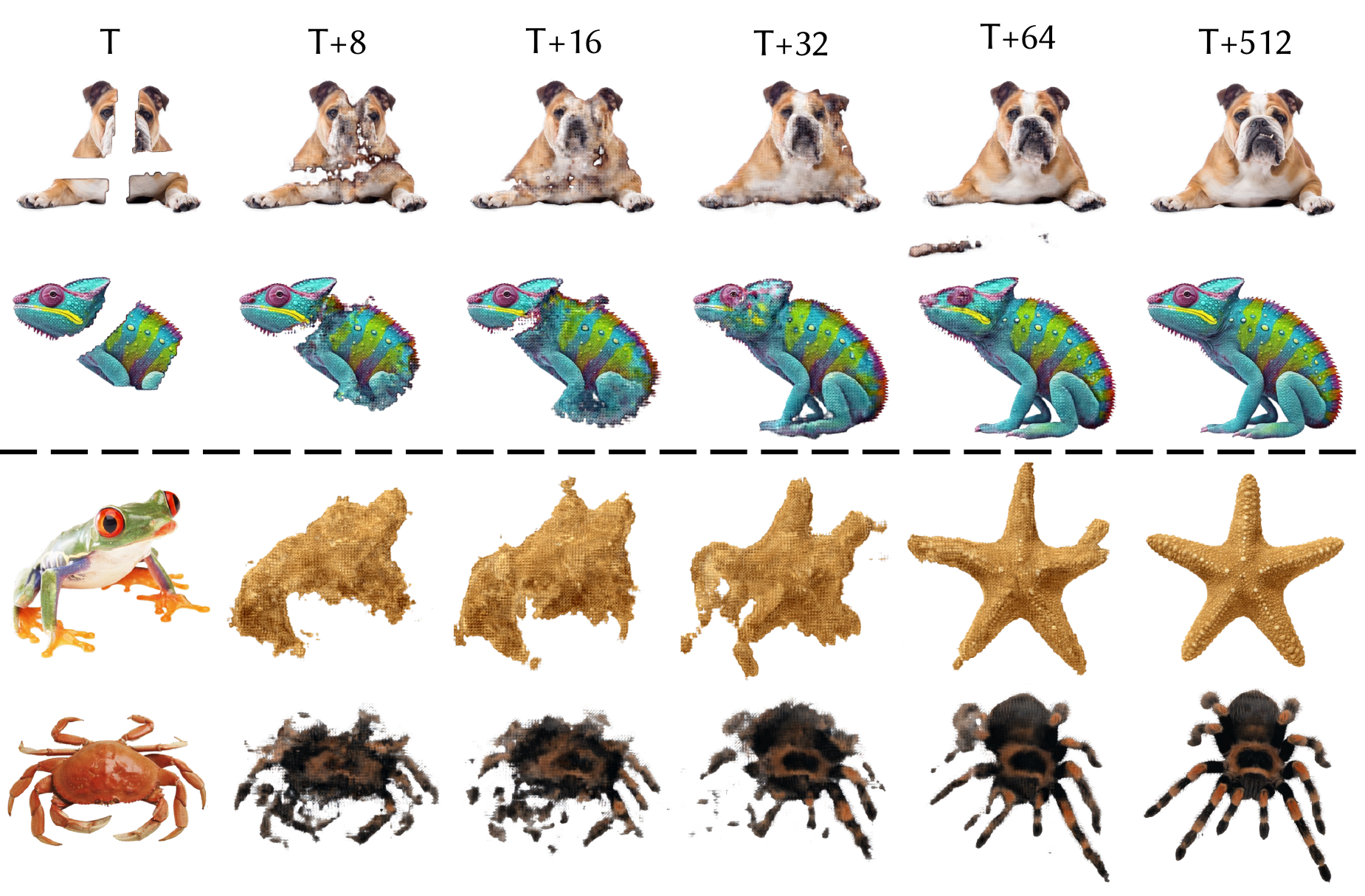}
    \vspace{-15pt}
    \caption{\textbf{Top:} Regeneration after damage. \textbf{Bottom:} Morphing between targets by switching model parameters.}
    \vspace{-5pt}
    \label{fig:morphology_regen_morph}
\end{figure}

\begin{figure*}[]
    \centering
    \includegraphics[width=\linewidth]{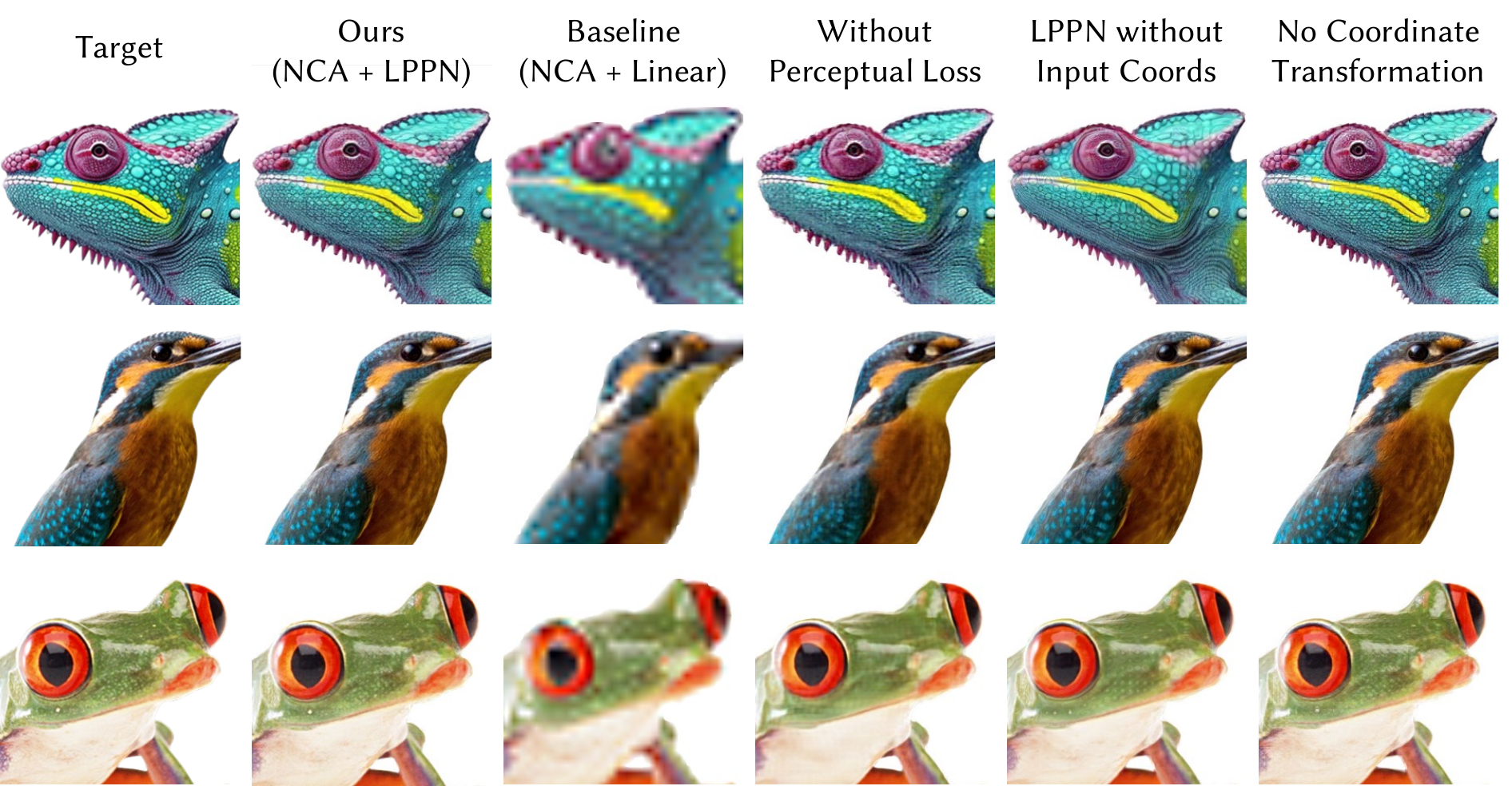}
    \caption{\rev{Ablations for "Growing a Morphology" experiment. Linear readout (no \ippn{}) suffers from low quality; removing LPIPS increases blur; removing coordinate input or $\sin/\cos$ encoding introduces blob-like or patch artifacts (best viewed digitally after zooming).}}

    \label{fig:morphology_ablation}
\end{figure*}

Figure~\ref{fig:morphology_regen_morph} highlights that the learned rule defines a stable self-organizing process: after severe damage the system re-grows missing regions and returns to the same attractor; swapping model parameters at test time produces a gradual morph from one converged morphology to another. 
\rev{Note that the morphing behavior is an emergent property and is not optimized explicitly; incorporating a dedicated morphing objective during training would yield more controllable and predictable transitions between attractors.}
Finally, Figure~\ref{fig:morphology_ablation} confirms the role of key components: removing \ippn{} or its coordinate input substantially degrades detail or introduces artifacts, and removing LPIPS or the $\sin/\cos$ encoding leads to blurry outputs or patch discontinuities.

\vspace{-4pt}
\subsection{PBR Texture Synthesis}
\label{sec:exp_pbr2d}

To synthesize 2D PBR textures the \ippn{} outputs a 9-channel field (albedo, normal, HRA). We apply a physics-based renderer to the synthesized maps only for visualizations and not during training.
This setting compresses $\approx 9\times1024^2$ target values into a model with $\approx $13k parameters.
For this experiment we use raw local coordinates for \ippn{} (no $\sin/\cos$), with similar observed performance.

\begin{figure*}[]
    \centering
    \includegraphics[width=\linewidth]{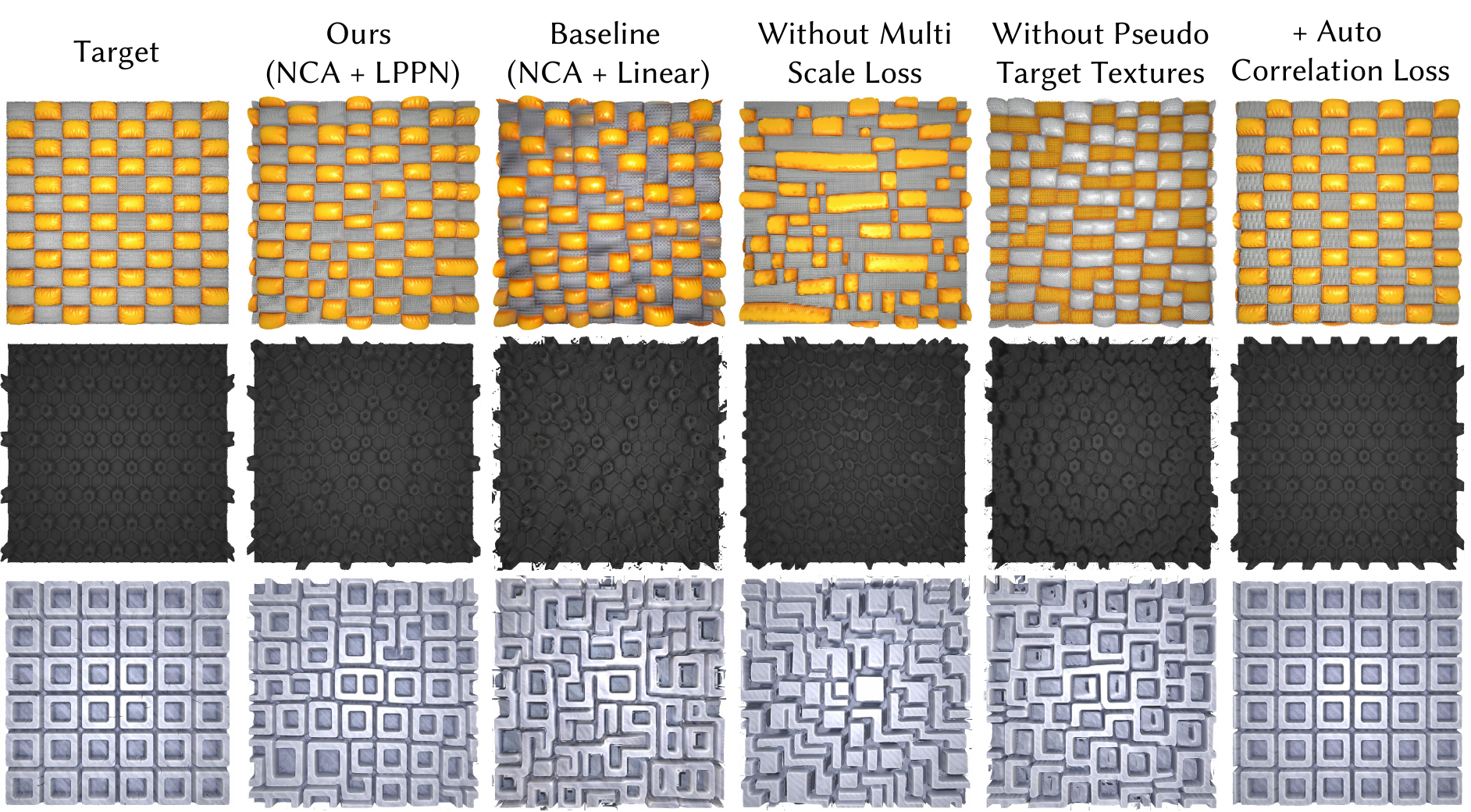}
    \caption{Results and ablations for "PBR Texture Synthesis" experiment. NCA+\ippn{} significantly improves the quality over the baseline; removing multi-scale loss hurts global coherence and removing pseudo targets breaks cross-map alignment; adding auto-correlation loss improves long-range geometric structure.}
    \label{fig:pbr2d_main}
\end{figure*}

Figure~\ref{fig:pbr2d_main} compares against the baseline and includes ablations on multi-scale supervision and pseudo targets, as well as the effect of the auto-correlation loss.
Multi-scale supervision improves global coherence, and pseudo targets improve cross-map alignment.
For textures with strong geometric regularity, the auto-correlation term improves long-range structure and shows that NCA dynamics can produce rigid, highly structured patterns under suitable long-range supervision despite their spontaneous dynamics and fluid-like nature~\citep{dynca}. We refer to the Appendix for additional results and visualization of all PBR maps.

\begin{figure}[]
\centering
\includegraphics[width=\linewidth]{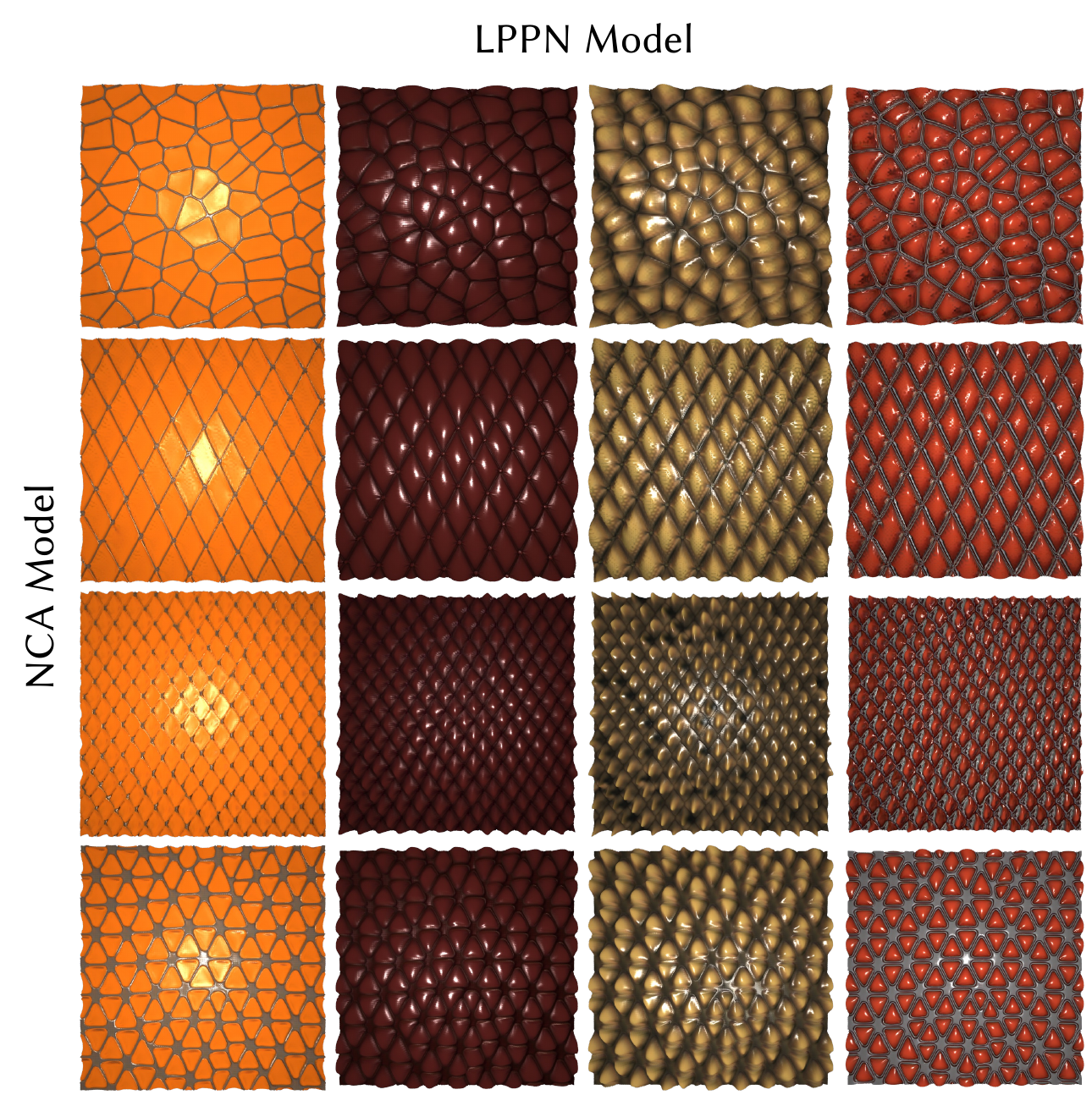}
\vspace{-10pt}
\caption{Swapping NCA and \ippn{} between trained models. Coarse layout follows the NCA, while fine detail follows the \ippn{}.On the diagonal, we show the original (unswapped) models.}
\vspace{-5pt}
\label{fig:pbr2d_decouple}
\end{figure}

\begin{figure}[]
\centering
\includegraphics[width=\linewidth, trim=0 0 0 0, clip]{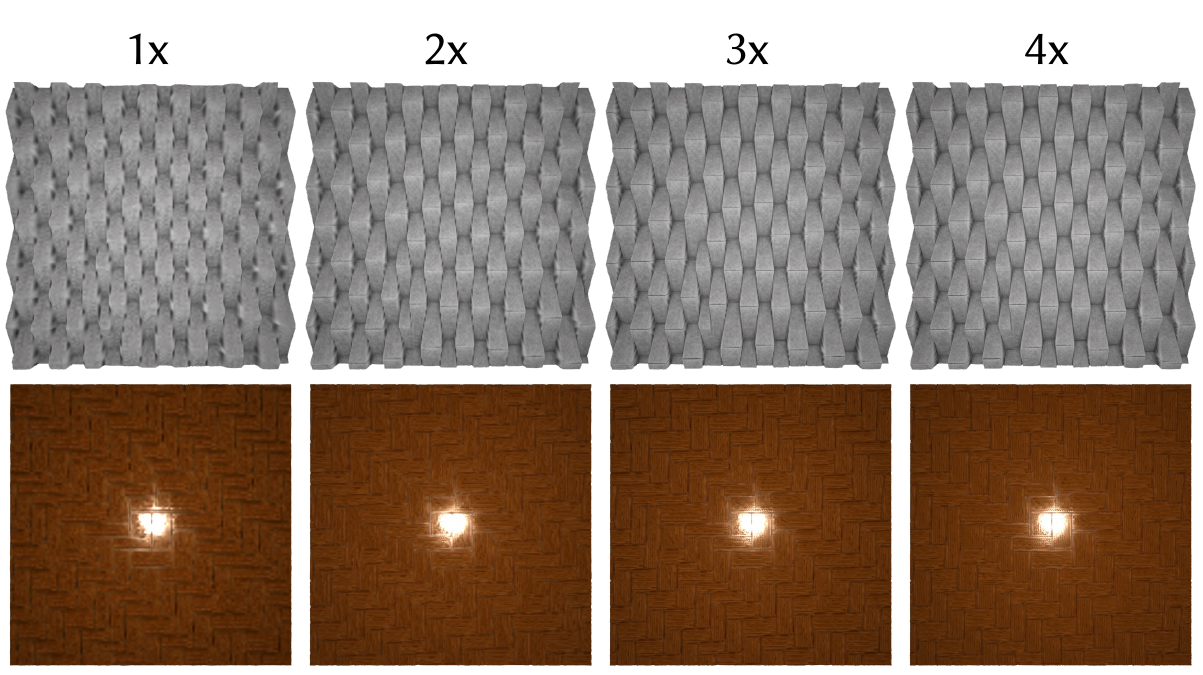}
\vspace{-10pt}
\caption{Increasing the \ippn{} evaluation resolution improves details and sharpness, enabling a compute--quality trade-off (training setting: 8$\times$).}\label{fig:pbr2d_lppn_scale}
\vspace{-10pt}
\end{figure}

Figure~\ref{fig:pbr2d_decouple} shows the effect of swapping NCA and \ippn{} between trained models, revealing a coarse-to-fine division of labor: the NCA determines the coarse geometric layout, while the \ippn{} contributes fine-scale appearance.
Figure~\ref{fig:pbr2d_lppn_scale} shows that rendering at resolutions different from training remains stable, and quality generally improves as the output resolution increases.
This provides a practical compute--quality knob: higher-resolution evaluation costs more but yields sharper and more detailed results.

\vspace{-3pt}
\subsection{Volumetric 3D Texture Synthesis}
\label{sec:exp_vol3d}

\begin{figure}[]
\centering
\includegraphics[width=\linewidth]{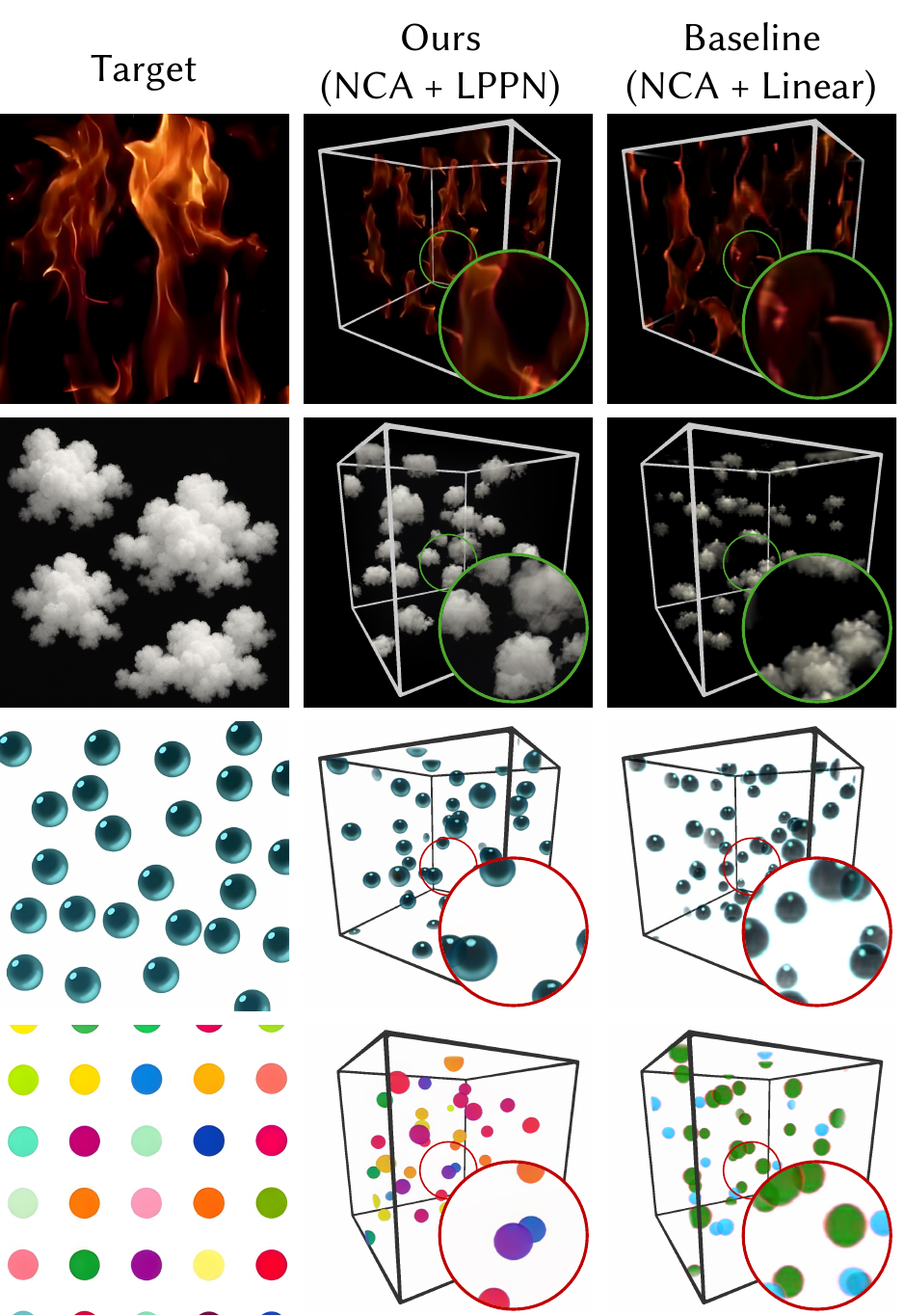}
\caption{3D texture synthesis results, and comparison to the baseline. }
\vspace{-5pt}
\label{fig:vol3d_results}
\end{figure}

We synthesize volumetric textures with an NCA on a $64^3$ grid representing a cube in $[-1,1]^3$.
To render a 2D image, we use NeRF-style volumetric rendering~\citep{nerf}: along each camera ray we sample points uniformly in the cube-ray intersection region, trilinearly interpolate the local NCA state, and query \ippn{} to obtain RGB color and a scalar density, followed by a \texttt{softplus} nonlinearity to ensure non-negative values for volumetric integration. 
We then composite samples along the ray using standard emission--absorption to produce the rendered image, which is supervised by the texture loss in Sec.~\ref{sec:texture-loss}.
We use an orthographic camera to reduce perspective distortion. 
We restrict viewpoints to respect the target's top--down and left--right symmetries.
Figure~\ref{fig:vol3d_results} shows that our model produces more coherent structure and substantially sharper details than the baseline.

\vspace{-7pt}
\subsection{Texture Synthesis on Meshes}
\label{sec:exp_mesh}

We adopt MeshNCA~\citep{meshnca}, where cells live on mesh vertices and perception is implemented via message passing over mesh connectivity using a spherical harmonics basis.
Following \citet{meshnca}, we train on an icosphere ($\sim$40k vertices) and apply the trained model to unseen meshes at test time without retraining.
In our hybrid setting, \ippn{} decodes locally averaged vertex states together with continuous surface coordinates to an RGB texture at arbitrary sampling points.

\begin{figure}[]
\centering
\includegraphics[width=\linewidth]{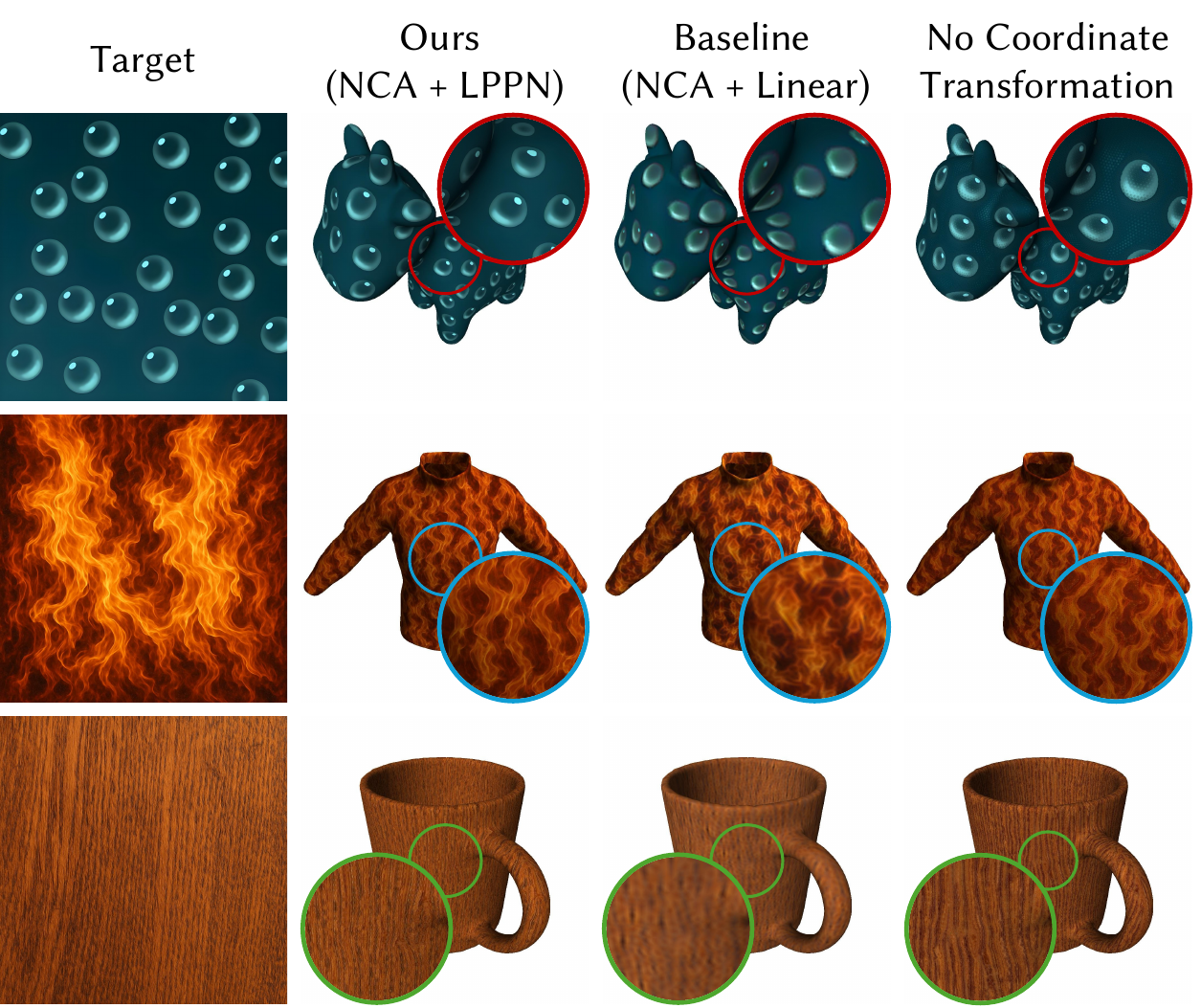}
\caption{Mesh texture synthesis results, comparison with the baseline, and an ablation showing the artifacts caused by removing the local coordinate transformation.}
\label{fig:mesh_results}
\end{figure}

For stable decoding across triangle boundaries, we apply a simple barycentric-coordinate preprocessing (sorting + remapping).
During training only (on the icosphere), we accelerate loss evaluation by replacing perspective rasterization with cached Lambert (equal-area) projections of random spherical patches.
These projections also reduce curvature and perspective distortions and allow us to precompute and reuse the barycentric coordinates and face indices for a fixed patch size, making training faster than repeatedly invoking a full rasterizer.
For the results shown in the paper and for testing on arbitrary meshes, we use a standard rasterizer with a perspective camera.
Figure~\ref{fig:mesh_results} shows improved sharpness and detail over the baseline and highlights the importance of our local coordinate transformation to prevent triangle-aligned artifacts.

\vspace{-7pt}
\subsection{Online Demo}

We provide an interactive WebGL demo at
\href{https://cells2pixels.github.io}{\textcolor{pinkl}{\nolinkurl{cells2pixels.github.io}}}
that runs our trained models fully on the GPU inside the browser (implemented with \texttt{SwissGL}). 
The core shader programs include: NCA updates, \ippn{} decoding/rendering, and interactive edits (erasing regions or inserting seeds). Additional controls let users adjust the simulation speed and the \ippn{} sampling scale.
The demo includes three modes corresponding to our main experiments: morphology growth (direct RGBA output), 2D PBR textures (9-channel maps visualized with a simple PBR shader), and 3D volumetric textures (NeRF-style ray-marching through the learned texture field).
All models are trained with an $8\times$ \ippn{} scale, but we default to smaller scales for real-time performance on edge devices (typically $4\times$ for morphology/PBR and $2\times$ for volumetric); in particular, volumetric rendering at $8\times$ may not run in real time without a strong GPU.

\vspace{-5pt}
\section{Conclusion}


We introduce a hybrid self-organizing framework that pairs Neural Cellular Automata with a lightweight \ippnfull{} to decouple NCA grid size from output resolution.
The NCA guides pattern formation through local updates on a coarse lattice, while the shared \ippn{} transforms a locally interpolated cell state together with continuous local coordinates into appearance attributes at any requested scale.
\rev{Our approach shares a high-level similarity with image super-resolution in that a learned decoder produces high-resolution outputs from a lower-resolution representation; however, our goal is not generic super-resolution from low-resolution pixels, but a decoder that is \emph{local}, \emph{lightweight}, and compatible with NCA evolution and its self-organizing properties. Moreover, our formulation extends naturally to non-image domains such as meshes and voxel grids.}
Our proposed task-specific loss functions designed for high-resolution outputs further improve the quality without exhausting memory and compute. We demonstrate that NCA, paired with our \ippn{}, is capable of generating detailed and high-resolution images and textures in real time. Our hybrid formulation enables NCA to scale to practical resolutions without losing its characteristic properties, such as robustness, controllability, and efficiency, paving the way for interactive, deployable self-organizing systems.

\textbf{Limitations.}
While \ippn{} substantially improves output resolution, it conditions only on intra-primitive coordinates and locally interpolated state (no access to cells outside the enclosing primitive), which can produce faint primitive-aligned patch artifacts.

\textbf{Future work.}
A natural next step is to extend our morphology setup from 2D objects to growing full 3D assets while preserving regeneration and controllability. On the optimization side, our texture losses already operate on patches; integrating this more tightly with \ippn{} evaluation (rendering only the queried regions used by the loss) could further reduce memory and enable training at even higher effective resolutions (e.g., $4$K and beyond). Beyond synthesis, the compactness of NCA+\ippn{} suggests applications in learned compression of images and textures. 
\rev{Finally, one could explore alternative formulations better suited to isotropic settings and to general polygons with varying numbers of vertices,
such as \emph{coordinate-free} \ippn{} variants that avoid explicit intra-primitive coordinates and instead condition on the interpolated cell state together with nearest-neighbor cell states.}

\newpage

\clearpage

\clearpage


\bibliographystyle{ACM-Reference-Format}
\bibliography{main}


\clearpage

\appendix

\appendix
\section{Additional Experiment Details}
\label{app:exp_details}

\subsection{Train and Inference Statistics}

\rev{
Table~\ref{tab:train_test_stats} summarizes training cost and inference throughput for each experiment (measured on an NVIDIA A100).
For the mesh setting, the reported render throughput depends heavily on the rasterization backend: using the Kaolin rasterizer end-to-end (rasterization + \ippn{} decoding) yields $\sim$6 steps/sec, while reusing a cached rasterization (fixed camera/visibility and barycentric interpolation weights) isolates the decoding cost and reaches $\sim$110 steps/sec.
We therefore report mesh throughput as \emph{6/110} to separate the rasterization bottleneck from the \ippn{} evaluation speed.
}

\begin{table}[t]
\centering
\caption{\rev{\textbf{Training and inference statistics} (NVIDIA A100).
For mesh inference, ``Steps/Sec (Render+\ippn{})'' is reported as \emph{6/110}: 6 with end-to-end Kaolin rasterization, 110 when rasterization results are cached and only \ippn{} decoding is evaluated.}}
\label{tab:train_test_stats}
\renewcommand{\arraystretch}{1.08}
\scriptsize
\resizebox{\columnwidth}{!}{
\begin{tabular}{llcccc}
\toprule
 & Setting & Growing & PBR-2D & Mesh & Vol-3D \\
\midrule
\multirow{4}{*}{Training} 
& Iterations & 20k & 3k & 3k & 3k \\
& 1k Iter. (minutes) & 5 & 5 & 5.5 & 35 \\
& Total Time (minutes) & 100 & 15 & 16 & 105 \\
& GPU Memory & 31 GB & 38 GB & 28 GB & 36 GB \\
\midrule
\multirow{4}{*}{Inference}
& Render Resolution & $768^2$ & $1024^2$ & $1024^2$ & $512^2$ \\
& Steps/Sec (NCA) & 2500 & 3700 & 1550 & 550 \\
& Steps/Sec (Render+\ippn{}) & 425 & 100 & 6/110 & 13 \\
& GPU Memory & 1.5 GB & 1.2 GB & 1.6 GB & 2.2 GB \\
\bottomrule
\end{tabular}
}
\vspace{-10pt}
\end{table}



\subsection{Common Training Recipe}
Across all experiments, we follow standard NCA training practices.
We maintain a checkpoint pool of intermediate NCA states to encourage long-term stability and expose the model to diverse rollouts; we use a pool size of 512 for all experiments and 1024 for morphology growth.
We use stochastic per-cell updates with probability $0.5$ and sample rollout lengths uniformly from the ranges in Table~\ref{tab:exp_configs}.
Every 32 training iterations, we reset the first batch element to the seed state (all zeros in all experiments except morphology growth, where the seed is a single active cell at the center). 
We include an overflow regularizer to keep states bounded,
$\lvert \State - \mathrm{clip}(\State,-1,1)\rvert$,
with weight 100 in all experiments, and normalize gradients of the NCA parameters for stability (we do not normalize \ippn{} gradients).
We use an initial learning rate of $10^{-3}$ with a step-decay schedule (factor $0.3$ every 1000 iterations), unless noted otherwise.
Following the grafting scheme of \citet{meshnca}, within each experiment we first train a single model from random initialization on an arbitrary target and use its converged NCA and \ippn{} weights to initialize all other runs in that experiment. \rev{Overall, we find that jointly training the NCA and \ippn{} is stable and requires no special treatment beyond the standard recipe used for vanilla NCA training.}

\subsection{Morphology Growth Training Schedule}
\label{app:morphology_training}
We train morphology growth for 20k iterations.
To reduce sensitivity to poor local minima, we periodically \emph{flush} the checkpoint pool by replacing all stored states with the seed and simultaneously restarting the learning-rate schedule; we do this every 4000 iterations.
In practice, perceptual quality saturates for many targets after $\sim$8000 iterations, while a subset continues to improve later in training.

\subsection{Auto-correlation Settings for PBR Textures}
\label{app:pbr_ac_details}
We enable the auto-correlation regularizer only for textures with strong geometric structure.
In our experiments, this term is more sensitive to hyperparameters than the base multi-scale OT loss: we typically use weights in the range 10--100 and, for some targets, extend training beyond the default 3k iterations (up to 20k) to reach convergence.
We apply the auto-correlation term only at the coarsest scale of the multi-scale texture loss, which we found sufficient in practice.

\subsection{Mesh Textures: Coordinate Transform and Training-Time Projection}
\label{app:mesh_exp_details}
\textbf{Barycentric coordinate preprocessing.}
On meshes, the natural local coordinates within a triangle are barycentric weights, but these are order-dependent because adjacent faces may index vertices differently.
To obtain a consistent coordinate field across triangle boundaries, we sort the barycentric coordinates at each sampling point (largest$\rightarrow$smallest) and then apply a lightweight remapping that balances the dynamic range across components.
We provide the full derivation of this remapping under a uniform-sampling assumption in Appendix~\ref{app:mesh_coords}.

\textbf{Cached Lambert projections for training.}
During training on the icosphere, we accelerate loss evaluation by replacing perspective rasterization with cached Lambert (equal-area) projections of random spherical patches.
This reduces curvature and perspective distortions and allows us to precompute barycentric coordinates and face indices for a fixed patch size, avoiding repeated rasterization during training.
For all paper results and for evaluation on arbitrary meshes, we revert to a standard rasterizer with a perspective camera.
More details are provided in Appendix~\ref{app:lambert_projection}.

\section{Mesh Local Coordinates: Sorting and Analytical CDF Remapping}
\label{app:mesh_coords}

For mesh texture synthesis, each sampling point $\Point$ lies in a triangle face and admits barycentric coordinates
$\boldsymbol{\lambda}(\Point)=(\lambda_1,\lambda_2,\lambda_3)$ with $\lambda_i\ge 0$ and $\sum_i \lambda_i=1$.
Directly feeding $\boldsymbol{\lambda}$ to the \ippn{} is problematic because adjacent faces may list their vertices in different orders, causing discontinuities across shared edges.
We therefore (i) sort the barycentric coordinates and (ii) remap each sorted component to obtain a comparable dynamic range across dimensions.

\subsection{Sorting for $C^0$ continuity across triangle edges}
\label{app:mesh_coords_sort}

Let
\begin{equation}
a=\max(\lambda_1,\lambda_2,\lambda_3), \quad
c=\min(\lambda_1,\lambda_2,\lambda_3), \quad
b = 1-a-c,
\end{equation}
so that $a\ge b\ge c$ are the sorted barycentric coordinates.
Along a shared edge, the two nonzero barycentric weights swap between the two incident faces, but sorting makes the ordered tuple $(a,b,c)$ identical on both sides; hence the local coordinate field becomes $C^0$ across triangle boundaries (non-differentiability occurs only on a measure-zero set where two weights are equal).

\subsection{Distribution of sorted barycentric coordinates under uniform sampling}
\label{app:mesh_coords_dist}

To equalize the dynamic range of $(a,b,c)$, we apply a monotone remapping based on their analytic distributions under uniform sampling within a triangle.
For an equilateral triangle, a uniformly sampled point has barycentric coordinates distributed uniformly over the 2-simplex, i.e.,
$(\lambda_1,\lambda_2,\lambda_3)\sim\mathrm{Dirichlet}(1,1,1)$.
Equivalently, the joint density of $(\lambda_1,\lambda_2)$ over $\{\lambda_1,\lambda_2\ge 0,\ \lambda_1+\lambda_2\le 1\}$ is constant, so probabilities reduce to area ratios.

We will repeatedly use the following identities (for $t\in[0,\tfrac12]$):
\begin{align}
\Pr(\lambda_1>t) &= (1-t)^2, \\
\Pr(\lambda_1>t,\ \lambda_2>t) &= (1-2t)^2, \\
\Pr(\lambda_1>t,\ \lambda_2>t,\ \lambda_3>t) &= (1-3t)^2 \quad (t\le \tfrac13).
\end{align}

\paragraph{Maximum $a=\max_i \lambda_i$.}
The support of $a$ is $[1/3,1]$.
For $t\in[1/3,1/2]$, we compute $\Pr(a\le t)$ via inclusion--exclusion:
\begin{equation}
\Pr(a\le t)
=
1 - 3\Pr(\lambda_1>t) + 3\Pr(\lambda_1>t,\lambda_2>t)
=
(3t-1)^2 .
\end{equation}
For $t\in[1/2,1]$, two components cannot both exceed $t$, hence
\begin{equation}
\Pr(a\le t)=1-3\Pr(\lambda_1>t)=1-3(1-t)^2.
\end{equation}
Differentiating yields a piecewise-linear density with a single peak at $t=\tfrac12$.

\paragraph{Minimum $c=\min_i \lambda_i$.}
The support of $c$ is $[0,1/3]$.
For $t\in[0,1/3]$, the event $\{c\ge t\}$ is the shrunken simplex
$\{\lambda_i\ge t,\ \sum_i\lambda_i=1\}$, obtained by the shift $\lambda_i'=\lambda_i-t$ and a uniform scaling by $(1-3t)$.
Since area scales quadratically, we obtain
\begin{equation}
\Pr(c\ge t)=(1-3t)^2
\quad\Rightarrow\quad
\Pr(c\le t)=1-(1-3t)^2.
\end{equation}

\paragraph{Middle component $b$.}
The support of $b$ is $[0,1/2]$.
The event $\{b>t\}$ is equivalent to ``at least two components exceed $t$''.
For $t\in[0,1/3]$, the triple intersection is possible, so
\begin{align}
\Pr(b>t)
& =
\Pr(\text{at least two } \lambda_i>t) \\
& =
3\Pr(\lambda_1>t,\lambda_2>t) - 2\Pr(\lambda_1>t,\lambda_2>t,\lambda_3>t) \\
& = 
3(1-2t)^2 - 2(1-3t)^2 .
\end{align}
For $t\in[1/3,1/2]$, the triple event is impossible and $\Pr(b>t)=3(1-2t)^2$.
This yields a continuous, piecewise-quadratic CDF (and a piecewise-linear density) with mode at $t=\tfrac13$.

\subsection{CDF remapping to uniform coordinates}
\label{app:mesh_coords_cdf}

The above derivations imply that each of the sorted coordinates follows an \emph{exact triangular distribution}:
\begin{equation}
a \sim \mathrm{Tri}\Big(\tfrac13,\,1,\,\tfrac12\Big),\qquad
b \sim \mathrm{Tri}\Big(0,\,\tfrac12,\,\tfrac13\Big),\qquad
c \sim \mathrm{Tri}\Big(0,\,\tfrac13,\,0\Big),
\end{equation}
where $\mathrm{Tri}(\ell,r,m)$ denotes a triangular distribution on $[\ell,r]$ with mode $m$.

We use the inverse CDF transform to map each component to a uniform-distribution coordinate.
For $x\sim\mathrm{Tri}(\ell,r,m)$, the CDF is
\begin{equation}
F_{\mathrm{Tri}}(x;\ell,r,m)=
\begin{cases}
0, & x<\ell,\\[2pt]
\dfrac{(x-\ell)^2}{(r-\ell)(m-\ell)}, & \ell\le x<m,\\[8pt]
1-\dfrac{(r-x)^2}{(r-\ell)(r-m)}, & m\le x\le r,\\[6pt]
1, & x>r.
\end{cases}
\label{eq:tri_cdf}
\end{equation}
We then rescale to $[-1,1]$:
\begin{equation}
u = 2\,F_{\mathrm{Tri}}(x;\ell,r,m) - 1.
\end{equation}
Applying this to $(a,b,c)$ yields a continuous local coordinate embedding with balanced dynamic range across the three components. The equilateral-triangle assumption is a close match for our training setup: icosphere faces are near-equilateral, and rasterized samples are approximately uniform within each face.
We therefore use the analytic CDFs above during training and inference.

\section{Lambert Patch Projection for Fast Icosphere Training}
\label{app:lambert_projection}

For mesh texture supervision, we need to evaluate a 2D image-space loss on signals defined over the sphere.
Using a perspective rasterizer during training introduces (i) perspective foreshortening and (ii) curvature-induced distortions.
Since MeshNCA is trained on a fixed icosphere~\citep{meshnca}, we can instead precompute a set of distortion-minimized surface patches and reuse them throughout training.

\subsection{Lambert azimuthal equal-area patches}
We parameterize a local spherical patch using the Lambert Azimuthal Equal-Area (LAEA) projection, which preserves area and therefore yields approximately uniform sampling density over the patch.
For each cached view, we sample a patch center (latitude/longitude) and define a regular $H\times W$ grid in the projection plane over a bounded radius.
We then apply the \emph{inverse} LAEA mapping to convert each grid point to a unit direction on the sphere.
(We use standard LAEA formulas via a projection library.)

\subsection{Caching barycentric coordinates and face indices}
Given unit directions $\{\mathbf{d}_{ij}\}$ from the sphere center, we intersect rays
$\mathbf{r}(t)=t\,\mathbf{d}_{ij}$ with the icosphere to obtain, for each pixel, the hit triangle index and barycentric coordinates within that triangle.
In practice we compute these intersections once for a fixed set of random patch centers and cache the results:
\begin{equation}
\texttt{faces} \in \mathbb{N}^{V\times H\times W\times 3},\qquad
\texttt{bary} \in \mathbb{R}^{V\times H\times W\times 3},
\end{equation}
where $V$ is the number of cached patches, each corresponding to a different viewing direction.
During training, we simply sample a subset of cached patches per iteration and use the stored $(\texttt{faces},\texttt{bary})$ to interpolate per-vertex NCA states to an image grid.
Because the cached tensors are constants, this is fully differentiable with respect to the vertex features (and \ippn{} parameters) while avoiding repeated rasterization.
We use the Lambert cache \emph{only during training on the icosphere} to speed up loss evaluation and reduce projection distortions. 
\rev{Note that MeshNCA does not use an explicit direction field but is axis-aware through the XYZ coordinate frame. During training, we restrict sampled camera viewpoints according to the symmetries of the target texture (e.g., isotropic, left-right symmetric, or asymmetric), helping the model learn consistent orientation of directional patterns.}
For all paper visualizations and for testing on general meshes, we revert to a standard rasterizer with a perspective camera, since the cached projection is sphere-specific and does not directly apply to arbitrary mesh geometry.

\section{Interactive Web Demo}
\label{app:demo}

\begin{figure*}[]
\centering
\includegraphics[width=\linewidth]{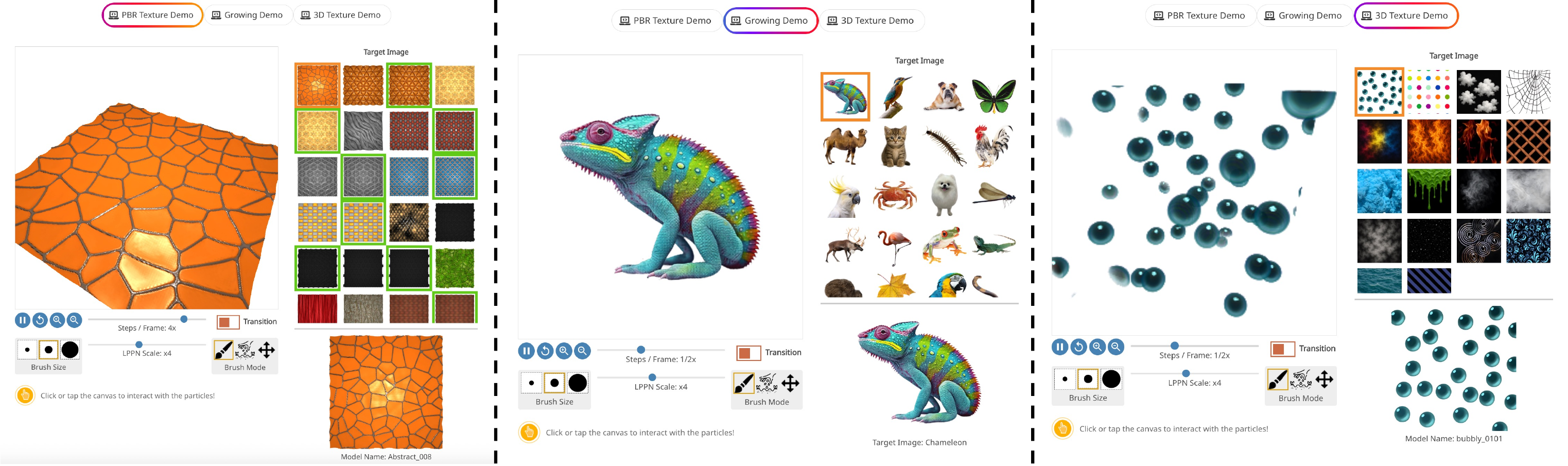}
\caption{Interactive web demo for three of our experiments: 2D PBR textures, morphology growth, and 3D volumetric textures.}
\label{fig:demo_ui}
\end{figure*}

We provide an interactive WebGL demo at
\href{https://cells2pixels.github.io}{\textcolor{pinkl}{\nolinkurl{cells2pixels.github.io}}}
that runs our trained models fully on the GPU inside the browser.
The demo is implemented with \texttt{SwissGL}, a lightweight wrapper around WebGL2 designed to minimize boilerplate for managing GLSL programs, textures, and framebuffers.

All three demo modes share the same high-level execution pattern.
Each animation frame alternates between (1) an NCA step shader that updates the cell state and (2) an \ippn{} shader that decodes the current state into the displayed output.
For efficiency, the NCA shader fuses perception (local neighborhood aggregation) and the two-layer update MLP into a single fragment shader program.
Model parameters are stored in buffers/textures or uniforms (loaded once per target), and the NCA state is stored in floating-point textures with ping--pong updates.

In all modes, users can directly perturb the NCA state by interacting with the canvas.
A brush tool either erases local state (setting cells/voxels to zero) or inserts a seed (in the morphology demo).
A speed control adjusts how many NCA steps are executed per rendered frame.
A scale slider controls the \ippn{} evaluation resolution: although all models are trained with an \ippn{} upscaling factor of $8\times$, the demo defaults to smaller scales to maintain real-time performance on mobile devices (we use $4\times$ for the morphology and PBR demos, and $2\times$ for the volumetric demo).

When the \textit{Transition} option is enabled, switching between targets does not reset the current state; instead, the newly selected model continues evolving the existing state.
This exposes cross-model morphing behavior (similar to the parameter-switch morphing shown in our experiments), and allows users to explore attractor changes interactively.

\subsection{Demo modes}
Figure~\ref{fig:demo_ui} shows the UI of our interactive web demo, including the three modes, target selection panel, and controls for  NCA steps per frame, \ippn{} scale, state editing, and model transitions.

\textbf{PBR textures (2D).}
The \ippn{} outputs $9$ channels corresponding to PBR maps (albedo, normal, height, roughness, and ambient occlusion).
For visualization, we render these maps with a simple PBR shader and a single point light source. The height channel is additionally used to displace the mesh geometry. The demo also includes an implementation of parallax occlusion mapping that is disabled to keep to UI simple. Targets highlighted with a green bounding box indicate models trained with auto-correlation supervision enabled.

\textbf{Morphology (2D).}
The \ippn{} directly produces an RGBA image from the current NCA state, which is displayed without additional shading.
Users can erase regions to test regeneration, or insert a seed to restart local growth.

\textbf{Volumetric textures (3D).}
We render a learned density field using a NeRF-style ray-marching shader.
For each pixel, we march along the camera ray through the canonical cube and repeatedly query the \ippn{} at sampled 3D locations (using local coordinates and interpolated NCA state), then composite color and density with standard front-to-back alpha integration.
This enables real-time exploration of animated volumetric textures, with interactive camera control and direct state editing.

\subsection{Targets}
Morphology targets are collected from a publicly available PNG repository, \href{https://pngimg.com}{https://pngimg.com}  and personal photos.
PBR texture targets use the dataset compiled by \citet{meshnca}.
Volumetric texture targets are drawn from three sources: texture images from DTD \citep{dtd}, the textures used in \citet{dynca}, and additional synthetic texture targets generated via text-to-image tools.

\section{Texture Expansion}
\label{app:texture_expansion}

\rev{
A key use case of texture synthesis is \emph{expansion}: producing a larger texture that remains consistent with a small exemplar.
NCAs support this naturally, since a trained update rule can be rolled out on larger lattices at test time.
Here we evaluate expansion for our model and study how the optional auto-correlation (AC) regularizer affects long-range structure.
}

\begin{figure*}[t]
    \centering
    \includegraphics[width=\linewidth]{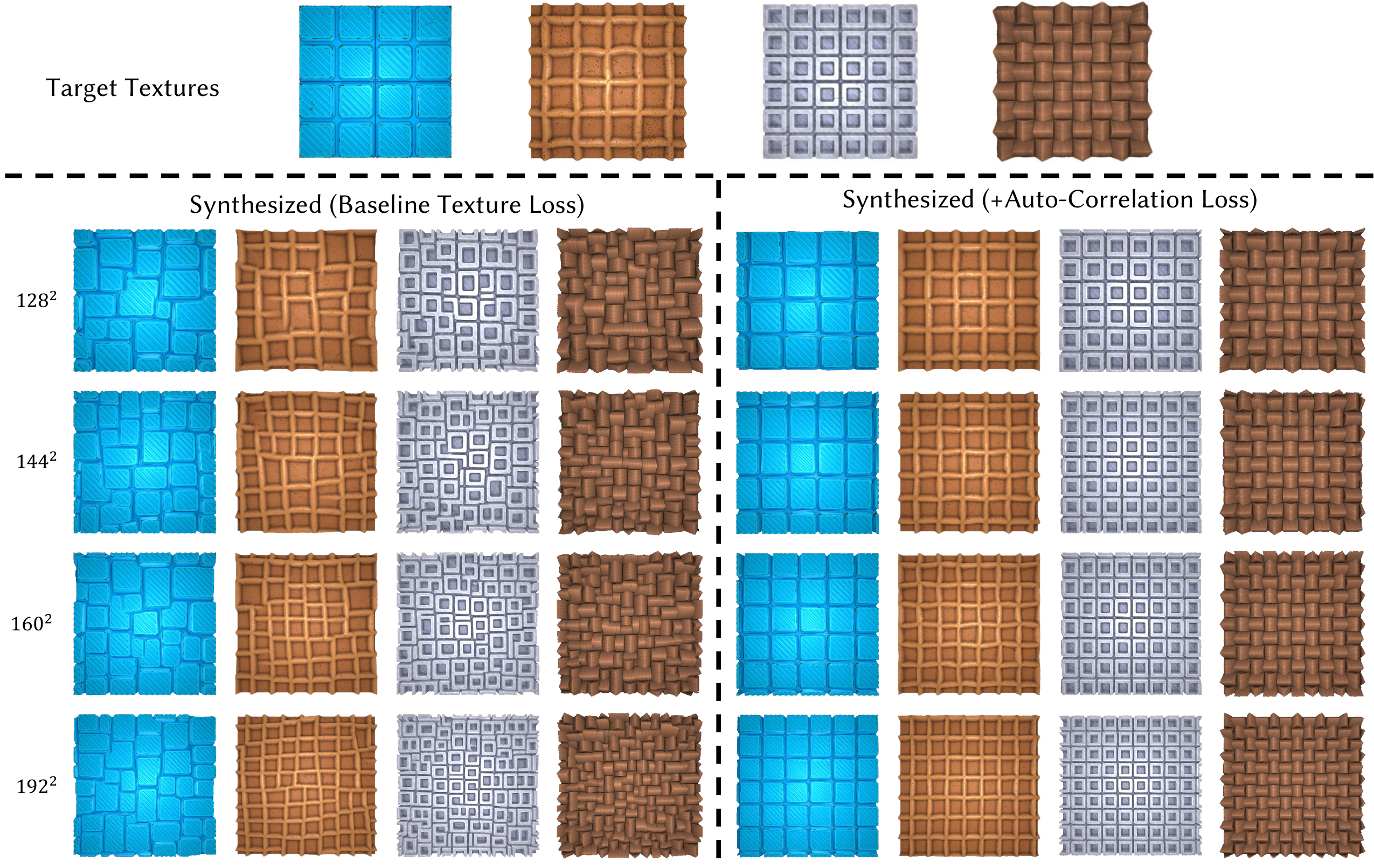}
    \caption{\rev{\textbf{Texture expansion across lattice sizes.}
    Models trained at $128^2$ are evaluated at $144^2$, $160^2$, and $192^2$.
    Left: baseline texture loss extends local appearance but loses long-range periodic alignment; right: adding auto-correlation preserves global lattice structure under expansion.}}
    \label{fig:texture_expansion}
\end{figure*}

\rev{
Figure~\ref{fig:texture_expansion} shows four targets with different underlying periodicities (roughly $4{\times}4$, $5{\times}5$, $6{\times}6$, and $8{\times}8$ tiles).
All models are trained on a $128^2$ NCA lattice, and then evaluated on larger lattices $144^2$, $160^2$, and $192^2$, corresponding to expansion factors $\tfrac{9}{8}$, $\tfrac{10}{8}$, and $\tfrac{12}{8}$.
With the baseline texture loss (left), the synthesized textures plausibly extend the local appearance, but global lattice alignment does not exist.
With AC supervision (right), the model better preserves the target's long-range periodic structure even under expansion.
For example, at $160^2$ ($\tfrac{10}{8}$ expansion), the $4{\times}4$ target expands to a clear $5{\times}5$ pattern, and the $8{\times}8$ target expands to $10{\times}10$.
}

\rev{
When the expected number of tiles under scaling is not an integer, the dynamics typically settle into one of the nearest integer lattice configurations.
For instance, the $6{\times}6$ target under $\tfrac{10}{8}$ expansion may converge to either a $7{\times}7$ or $8{\times}8$ tiling, depending on stochastic rollout.
}

\section{More Results}

\begin{figure*}[]
    \centering
    \includegraphics[width=\linewidth]{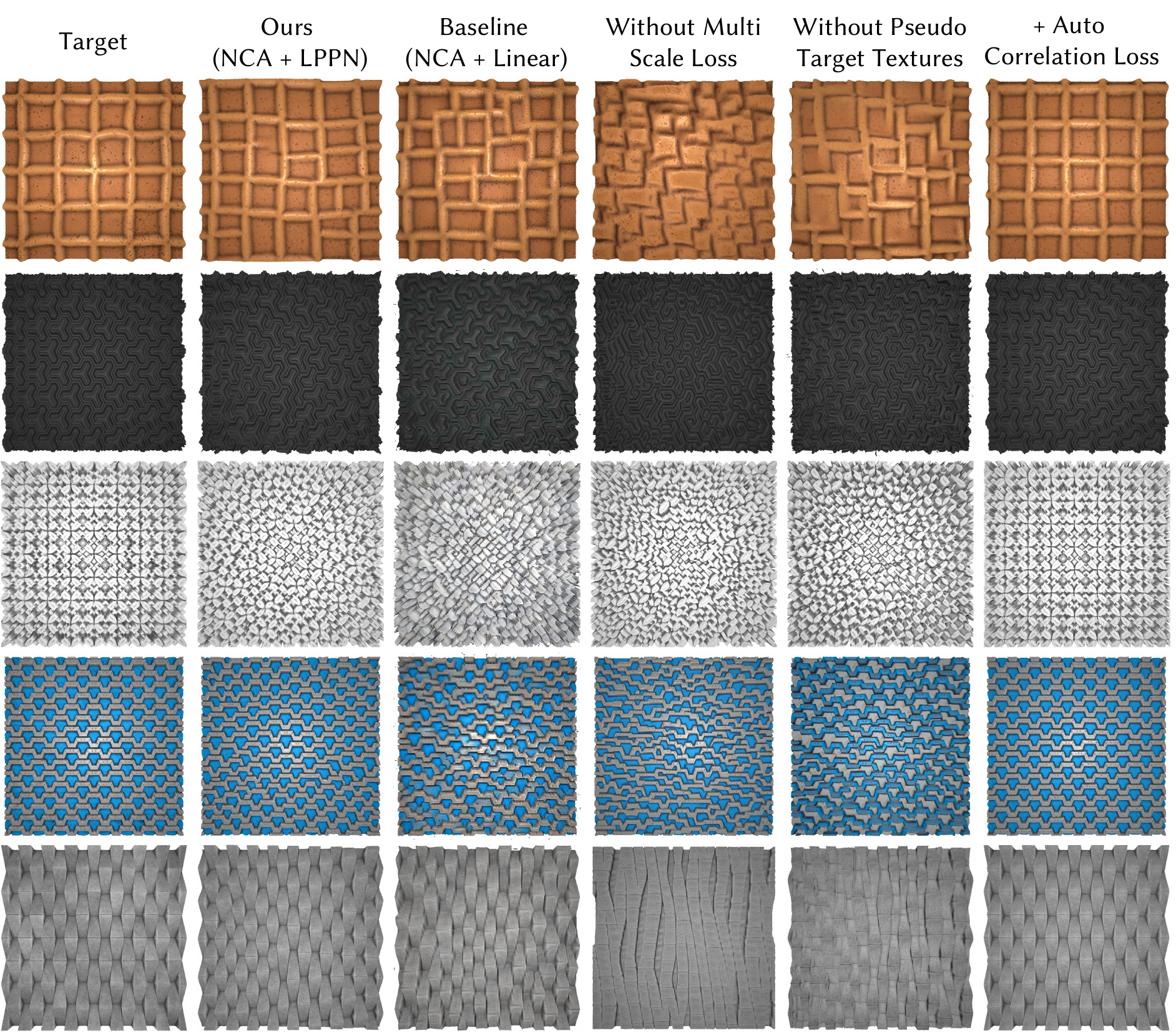}
    \caption{Results and ablations for "PBR Texture Synthesis" experiment. NCA+\ippn{} significantly improves the quality over the baseline; removing multi-scale loss hurts global coherence and removing pseudo targets breaks cross-map alignment; adding auto-correlation loss improves long-range geometric structure.}
    \label{fig:pbr2d_more}
\end{figure*}

\begin{figure*}[]
\centering
\includegraphics[width=0.85\linewidth]{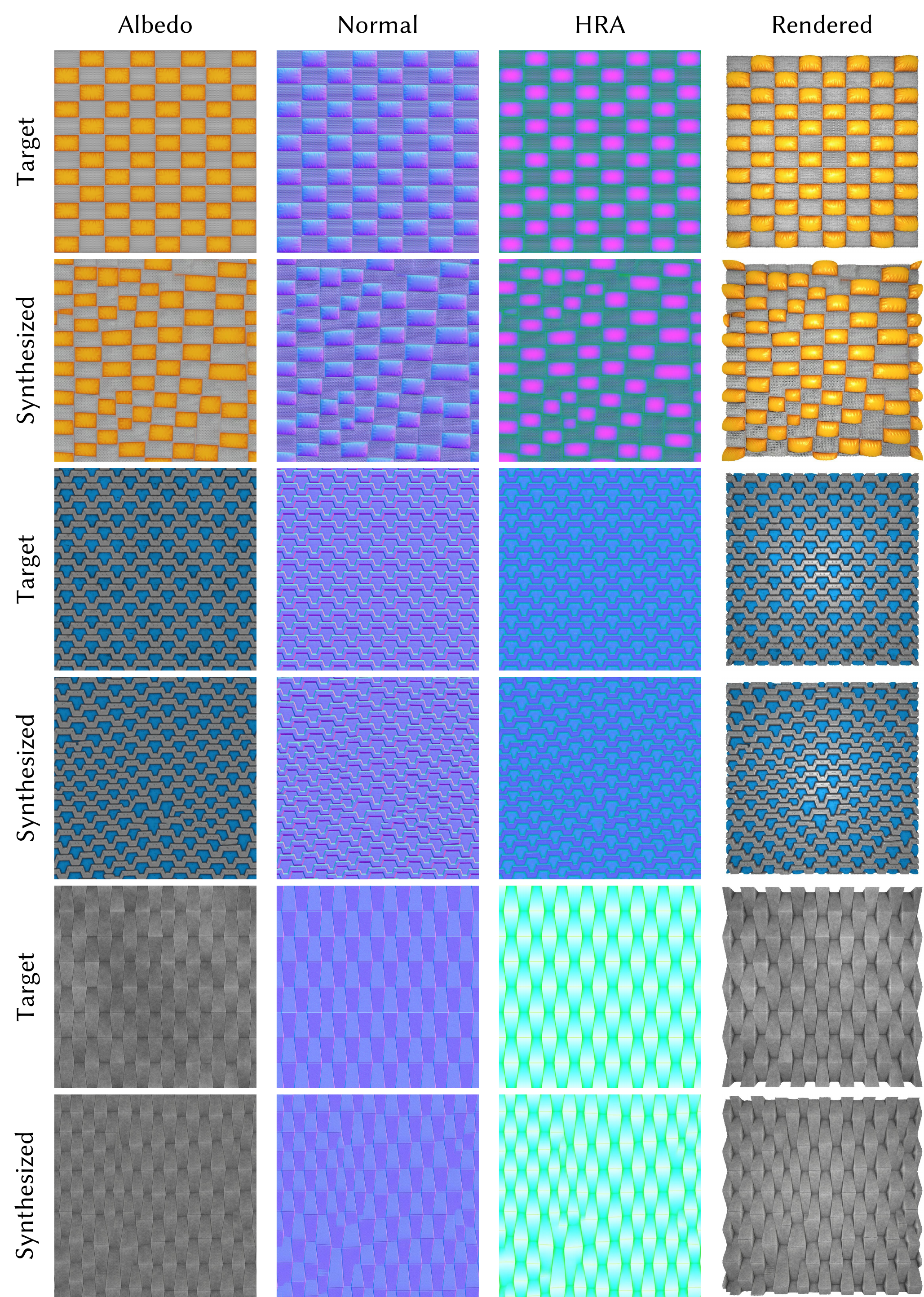}
\caption{Visualization of all texture maps (target and synthesized) for the PBR texture experiment.}
\label{fig:2dpbr_all1}
\end{figure*}
\begin{figure*}[]
\centering
\includegraphics[width=0.85\linewidth]{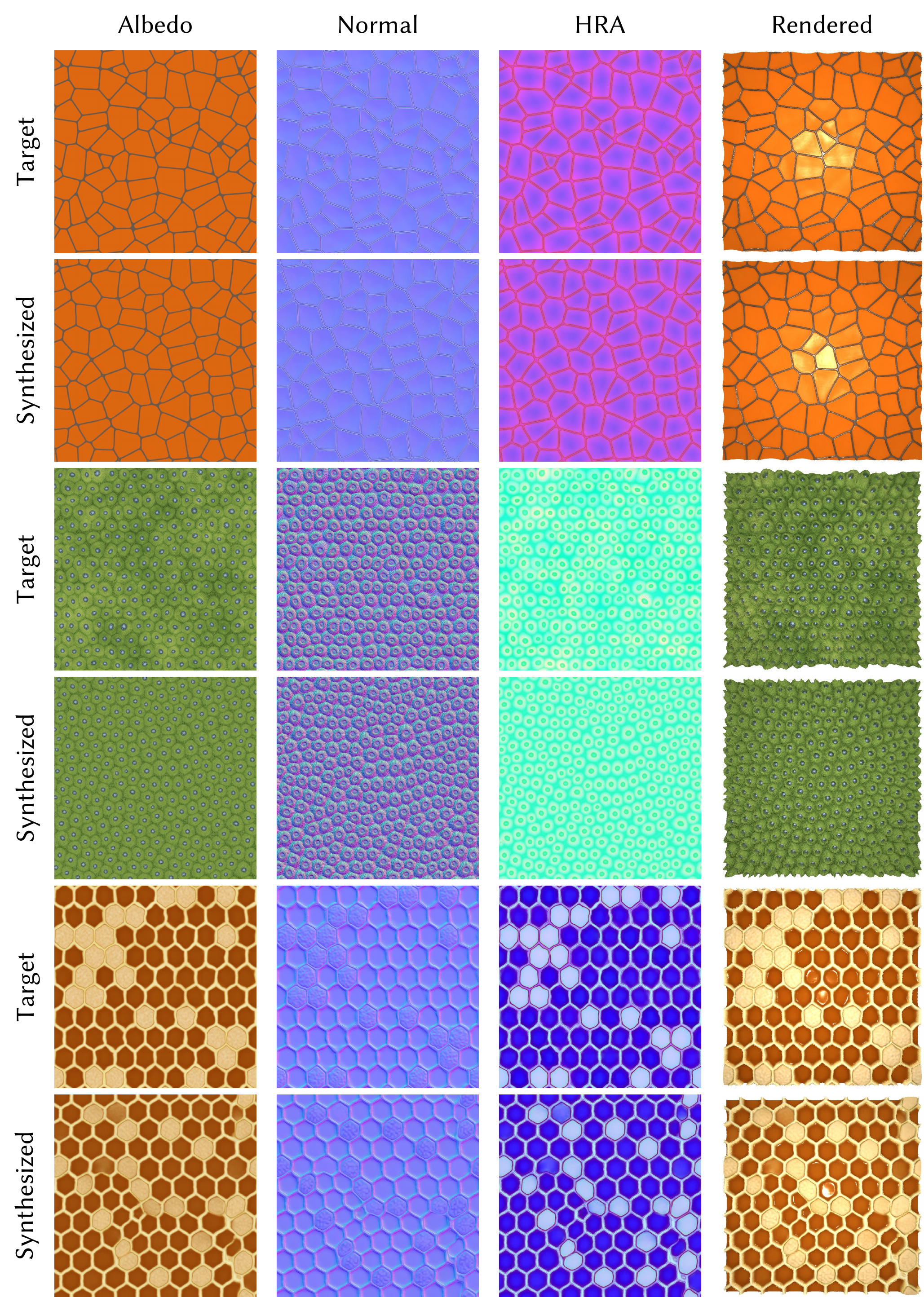}
\caption{Visualization of all texture maps (target and synthesized) for the PBR texture experiment.}
\label{fig:2dpbr_all2}
\end{figure*}
\begin{figure*}[]
\centering
\includegraphics[width=0.85\linewidth]{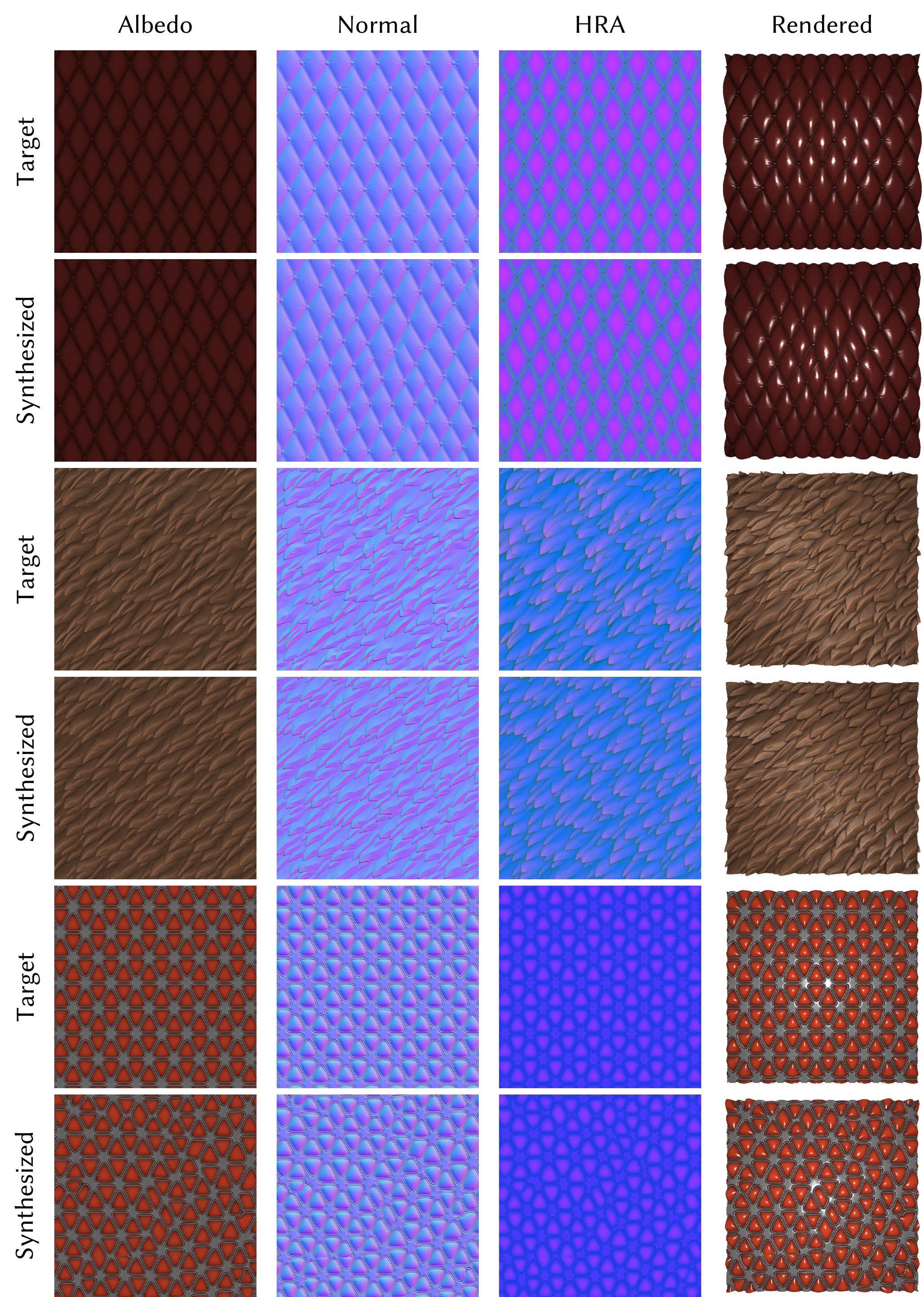}
\caption{Visualization of all texture maps (target and synthesized) for the PBR texture experiment.}
\label{fig:2dpbr_all3}
\end{figure*}

Figure~\ref{fig:pbr2d_more} shows more results for the PBR texture synthesis experiments.
We provide additional qualitative results for the 2D PBR texture experiment by visualizing the full set of target and synthesized texture maps, along with the corresponding renderings in Figures~\ref{fig:2dpbr_all1}-\ref{fig:2dpbr_all2}-\ref{fig:2dpbr_all3}.
Each target consists of 9 channels grouped into three maps: \emph{Albedo} (RGB), \emph{Normal} (RGB), and \emph{HRA} (height, roughness, ambient occlusion).
For visualization, we render the texture maps using a simple PBR shader implemented in PyTorch, including parallax occlusion mapping to account for the height channel.

Overall, the synthesized maps closely match the targets, and the rendered appearance remains faithful despite training only on per-map supervision (i.e., without any loss on the final rendered image).
In some cases, we observe mild patch-like artifacts in regions with strong specular response, where small local deviations in the predicted maps can be amplified by the renderer; this is consistent with the absence of direct supervision on the composed rendering.
We leave incorporating a rendering-aware objective (and related strategies for reducing such artifacts) to future work.

\end{document}